\documentclass[10pt,twocolumn,letterpaper]{article}

\usepackage[pagenumbers]{cvpr} 

\usepackage[dvipsnames]{xcolor}

\usepackage{times}
\usepackage{epsfig}
\usepackage{graphicx}
\usepackage{amsmath}
\usepackage{amssymb}
\usepackage{mathtools}
\usepackage{multirow}
\usepackage{enumitem}
\usepackage{caption}
\usepackage{subcaption}
\usepackage{array}
\usepackage{colortbl}
\usepackage{booktabs}
\usepackage{bbm}
\usepackage{multicol}
\usepackage{makecell}
\usepackage{xcolor}
\usepackage{xspace}
\usepackage{float}
\usepackage{color}
\usepackage{siunitx}
\usepackage{pifont}
\usepackage{marvosym}

\definecolor{linkColor}{rgb}{0.18,0.39,0.62}
\usepackage[pagebackref=true,breaklinks=true,colorlinks,citecolor=linkColor]{hyperref}

\def\modelname{InternVL\xspace}

\def\confName{CVPR}
\def\confYear{2024}

\title{\modelname: Scaling up Vision Foundation Models and Aligning \\ for Generic Visual-Linguistic Tasks}

\author{
    Zhe Chen$^{2,1\dagger}$,
    Jiannan Wu$^{3,1\dagger}$,
    Wenhai Wang$^{1,4}$,
    Weijie Su$^{6,1\dagger}$,
    Guo Chen$^{2,1\dagger}$, 
    Sen Xing$^{5}$,
    Muyan Zhong$^{5}$, \\
    Qinglong Zhang$^{1}$,
    Xizhou Zhu$^{5,7,1}$, 
    Lewei Lu$^{7,1}$,
    Bin Li$^6$,
    Ping Luo$^3$,
    Tong Lu$^{2}$,
    Yu Qiao$^{1}$,
    Jifeng Dai$^{5,1}$\textsuperscript{\Letter}  \\ 
    $^1$OpenGVLab, Shanghai AI Laboratory~~~
    $^2$Nanjing University\\
    $^3$The University of Hong Kong~~~
    $^4$The Chinese University of Hong Kong~~~
    $^5$Tsinghua University \\
    $^6$University of Science and Technology of China~~~
    $^7$SenseTime Research \\
	{\small \url{https://github.com/OpenGVLab/InternVL}} \\
}

\definecolor{baselinecolor}{gray}{.9}
\newcommand\blfootnote[1]{%
\begingroup
\renewcommand\thefootnote{}\footnote{#1}%
\addtocounter{footnote}{-1}%
\endgroup
}

\begin{document}

\twocolumn[{%
\maketitle
\vspace{-8mm}
\begin{figure}[H]
\hsize=\textwidth
\centering
\vspace{-1em}
\includegraphics[width=0.95\textwidth]{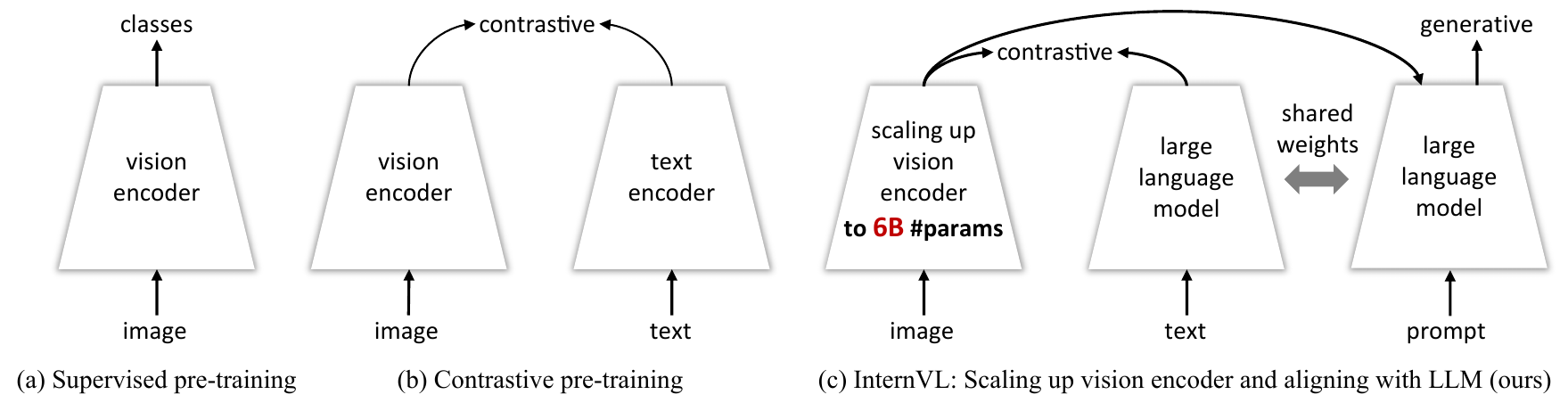}
\vspace{-0.5em}
\caption{\textbf{Comparisons of different vision and vision-language foundation models.} (a) indicates the traditional vision foundation model, \eg ResNet~\cite{he2016deep} pre-trained on classification tasks. (b) represents the vision-language foundation models, \eg CLIP~\cite{radford2021clip} pre-trained on image-text pairs. (c) is our InternVL, which presents a workable way to align the large-scale vision foundation model (\ie, InternViT-6B) with the large language model and is versatile for both contrastive and generative tasks. 
}
\label{fig:moti}
\end{figure}
}]

\begin{abstract}
\vspace{-1.3em}
\blfootnote{$\dagger$ This work is done when they are interns at Shanghai AI Laboratory;
\Letter~corresponding author (daijifeng@tsinghua.edu.cn)}
The exponential growth of large language models (LLMs) has opened up numerous possibilities for multi-modal AGI systems. 
However, the progress in vision and vision-language foundation models, which are also critical elements of multi-modal AGI, has not kept pace with LLMs. 
In this work, we design a large-scale vision-language foundation model (\modelname), which scales up the vision foundation model to 6 billion parameters and progressively aligns it with the LLM, using web-scale image-text data from various sources. 
This model can be broadly applied to and achieve state-of-the-art performance on 32 generic visual-linguistic benchmarks including visual perception tasks such as image-level or pixel-level recognition, vision-language tasks such as zero-shot image/video classification, zero-shot image/video-text retrieval, and link with LLMs to create multi-modal dialogue systems. It has powerful visual capabilities and can be a good alternative to the ViT-22B.
We hope that our research could contribute to the development of multi-modal large models.

\end{abstract}

\section{Introduction}

Large language models (LLMs) largely promote the development of artificial general intelligence (AGI) systems with their impressive capabilities in open-world language tasks, and their model scale and performance are still increasing at a fast pace. 
Vision large language models (VLLMs) \cite{wang2023visionllm, bai2023qwenvl, liu2023llava, zhu2023minigpt4, chen2023shikra, peng2023kosmos2, instructblip, alayrac2022flamingo, chen2022pali}, which leverage LLMs, have also achieved significant breakthroughs, enabling sophisticated vision-language dialogues and interactions. 
However, the progress of vision and vision-language foundation models, which are also crucial for VLLMs, has lagged behind the rapid growth of LLMs.

\begin{figure*}[t]
    \centering
    \includegraphics[width=1\textwidth]{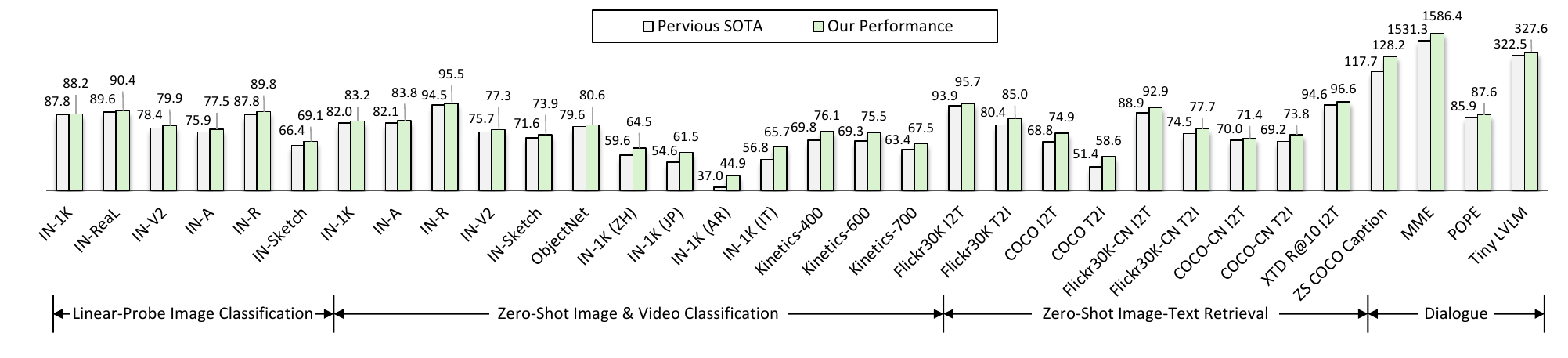}
    \caption{\textbf{Comparison results on various generic visual-linguistic tasks}, including image classification, video classification, image-text retrieval, image captioning, and multi-modal dialogue. The proposed InternVL achieves the best performance on all these tasks. Note that only the models trained on public data are included. 
    ``IN" is an abbreviation for ImageNet~\cite{deng2009imagenet}.
    } 
    \label{fig:sota_result}
\end{figure*}

To bridge vision models with LLMs, existing VLLMs \cite{li2023blip2, zhu2023minigpt4, bai2023qwenvl, zhang2023internlmxcomposer, sun2023emu} commonly employ lightweight ``glue" layers, such as QFormer \cite{li2023blip2} or linear projection \cite{liu2023llava}, to align features of vision and language models. 
Such alignment contains several limitations:
(1)~\emph{Disparity in parameter scales.}
The large LLMs~\cite{fedus2022switch} now boosts up to 1000 billion parameters, while the widely-used vision encoders of VLLMs are still around one billion.
This gap may lead to the under-use of LLM's capacity.
(2)~\emph{Inconsistent representation.}
Vision models, trained on pure-vision data or aligned with the BERT series \cite{devlin2018bert,liu2019roberta,jia2021scaling}, often exhibit representation inconsistencies with LLMs.
(3)~\emph{Inefficient connection.} The ``glue'' layers are usually lightweight and randomly initialized, which may not capture the rich cross-modal interactions and dependencies that are crucial for multi-modal understanding and generation.

These limitations reveal a large gap in both parameter scale and feature representation ability between the vision encoder and the LLM.
To bridge this gap, \emph{our inspiration lies in elevating the vision encoder to align with the parameter scale of the LLM and subsequently harmonizing their representations.}
However, the training of such large-scale models necessitates a vast amount of image-text data obtained from the Internet. The significant heterogeneity and quality variations within this data pose considerable challenges to the training process.
To enhance the efficacy of the training, generative supervision is considered as a complementary approach to contrastive learning, as depicted in Figure~\ref{fig:moti}. This strategy aims to provide additional guidance to the model during training.
Yet, the suitability of low-quality data for generative training remains a concern.
Besides, how to effectively represent the users' commands and align the representations between the vision encoder and LLM is another open question.

To address these issues, we formulate the \emph{InternVL, a large-scale vision-language foundation model, which aligns the representation of the scaled-up vision encoder with the LLM and achieves state-of-the-art performance on various visual and vision-language tasks.}
As shown in Figure \ref{fig:moti} (c), \modelname has three key designs: (1)  \emph{Parameter-balanced vision and language components}: It includes a vision encoder scaled up to 6 billion parameters and an LLM middleware with 8 billion parameters, where the middleware functions as a substantial ``glue'' layer to reorganize visual features based on user commands. 
Unlike prior vision-only (Figure \ref{fig:moti} (a)) or dual-tower (Figure \ref{fig:moti} (b)) structures, our vision encoder and middleware offer flexible combinations for both contrastive and generative tasks.
(2) \emph{Consistent representations}: To maintain the consistency of representations between the vision encoder and LLM, we employ a pre-trained multilingual LLaMA~\cite{cui2023chinesellama}, to initialize the middleware and align the vision encoder with it.
(3) \emph{Progressive image-text alignment}: We leverage image-text data from diverse sources, ensuring training stability through a progressive alignment strategy. This strategy initiates contrastive learning on large-scale noisy image-text data and subsequently transitions to generative learning on fine-grained data. This approach ensures a consistent enhancement of model performance and task scope.

These designs endow our model with several advantages:
(1) \emph{Versatile.} It functions as a standalone vision encoder for perception tasks, or collaborates with the language middleware for vision-language tasks and multi-modal dialogue systems. The language middleware bridges the gap between the vision encoder and the LLM decoder.
(2) \emph{Strong.} By leveraging the training strategy, large-scale parameters, and web-scale data, our model has a powerful representation that helps to achieve state-of-the-art results on various vision and vision-language tasks, as shown in Figure~\ref{fig:sota_result}. 
(3) \emph{LLM-friendly.} Due to the aligned feature space with LLMs, our model can smoothly integrate with existing LLMs, such as LLaMA series~\cite{touvron2023llama, touvron2023llama2}, Vicuna~\cite{zheng2023vicuna}, and InternLM~\cite{2023internlm}. These features distinguish our model from the previous approaches and establish a leading vision-language foundation model for various applications.

In summary, our contribution has three folds:

(1) We present a large-scale vision-language foundation model---\modelname, which aligns the large-scale vision encoder with LLMs for the first time. The model demonstrates strong performance on a wide range of generic visual-linguistic tasks, including visual perception tasks, vision-language tasks, and multi-modal dialogue.

(2) We introduce a progressive image-text alignment strategy for the efficient training of large-scale vision-language foundation models. This strategy maximizes the utilization of web-scale noisy image-text data for contrastive learning and fine-grained, high-quality data for generative learning.

(3) We extensively compare the proposed model with the current state-of-the-art vision foundation models and VLLMs. 
The results indicate that \modelname achieves leading performance on a broad range of generic visual-linguistic tasks, including image classification (ImageNet), semantic segmentation (ADE20K), video classification (Kinetics), image-text retrieval (Flickr30K \& COCO), video-text retrieval (MSR-VTT), and image captioning (COCO \& Flickr30K \& NoCaps). Meanwhile, it is also effective for multi-modal dialogue (MME \& POPE \& Tiny LVLM).

\section{Related Work}
\label{sec:related_work}

\begin{figure*}[t!]
    \centering
    \includegraphics[width=0.9\textwidth]{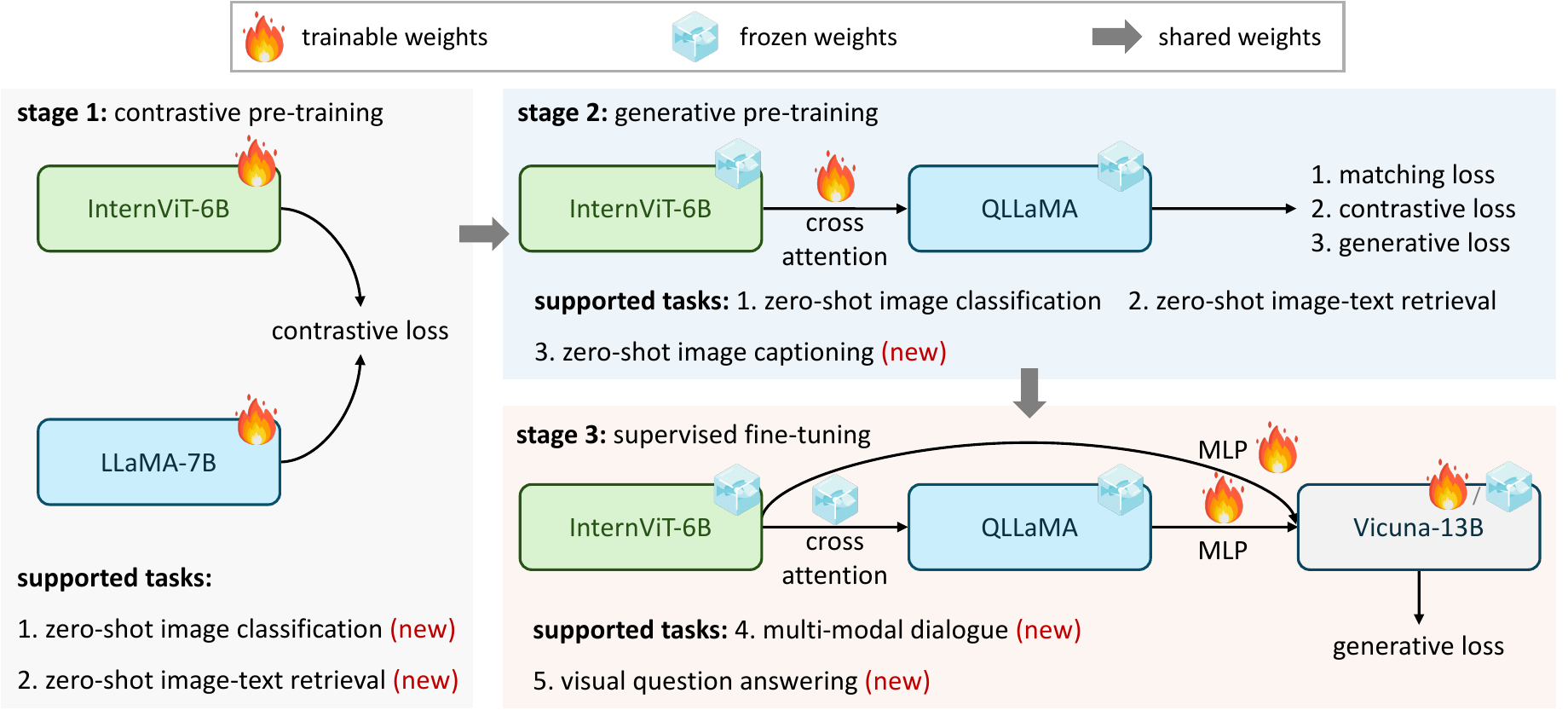}
    \caption{\textbf{The training strategy of the proposed \modelname model. }
    It consists of three progressive stages, including vision-language contrastive training, vision-language generative training, and supervised fine-tuning. These stages effectively leverage public data from diverse sources, ranging from noisy image-text pairs on the web to high-quality caption, VQA, and multi-modal dialogue datasets.
    } 
    \label{fig:training}
\end{figure*}

\subsection{Vision Foundation Models}

The past decade has witnessed significant development in foundation models within the field of computer vision. Starting with the pioneering AlexNet~\cite{krizhevsky2012imagenet}, a variety of convolutional neural networks (CNNs) have emerged, continuously refreshing the ImageNet benchmark~\cite{he2016deep, liu2022convnet, wang2023internimage, xie2017aggregated, iandola2014densenet, ding2021repvgg, dai2017deformable, hu2018senet}. In particular, the introduction of residual connections~\cite{he2016deep} effectively addressed the problem of vanishing gradients.
This breakthrough led to an era of ``big \& deep" neural networks, 
signifying that, with adequate training and data, larger and deeper models can achieve better performance. In other words, scaling up matters.

In recent years, ViT~\cite{dosovitskiy2020image} has opened up new possibilities for network architectures in the computer vision field. 
ViT and its variants \cite{wang2021pyramid,wang2021pvtv2,zhang2021rest,zhang2022rest,liu2021swin,dehghani2023vit22b,fang2022eva,radford2021clip, chen2022vitadapter, cai2022reversible} have significantly increased their capacity and excelled in various important visual tasks.
In the LLM era, these vision foundation models often connect with LLMs through some lightweight ``glue'' layers~\cite{liu2023llava, li2022blip, zhu2023minigpt4}.
However, a gap exists as these models primarily derive from visual-only datasets like ImageNet \cite{deng2009imagenet} or JFT \cite{zhai2022scaling}, or are aligned with the BERT series \cite{devlin2018bert, liu2019roberta, jia2021scaling} using image-text pairs, lacking direct alignment with LLMs. 
Additionally, the prevalent vision models employed to connect with LLMs are still limited to around 1 billion parameters~\cite{fang2022eva,openclip}, which also constrains the performance of VLLMs.

\subsection{Large Language Models}
Large language models (LLMs) have revolutionized the field of artificial intelligence, enabling natural language processing tasks that were previously thought exclusive to humans \cite{wei2022chain,openai2023gpt4,touvron2023llama}. The emergence of GPT-3~\cite{wei2022chain} brought a significant leap in capabilities, particularly in few-shot and zero-shot learning, highlighting the immense potential of LLMs. This promise was further realized with the advancements of ChatGPT and GPT-4~\cite{openai2023gpt4}.
The progress in the field has been further accelerated by the emergence of open-source LLMs, including the LLaMA series~\cite{touvron2023llama,touvron2023llama2}, Vicuna \cite{zheng2023vicuna}, InternLM \cite{2023internlm}, MOSS \cite{sun2023moss}, ChatGLM \cite{du2022glm}, Qwen~\cite{qwen}, Baichuan \cite{baichuan2023baichuan2}, and Falcon \cite{refinedweb}, among others~\cite{taori2023alpaca, wei2023skywork, cui2023chinesellama}. 
However, in real scenarios, interactions are not limited to natural language.
The vision modality can bring additional information, which means more possibilities. 
Therefore, exploring how to utilize the excellent capabilities of LLMs for multi-modal interactions is poised to become the next research trend.

\subsection{Vision Large Language Models}

Recent advancements have seen the creation of vision large language models (VLLMs) \cite{zhang2023llama-adapter, zhang2023internlmxcomposer, zhang2023gpt4roi, wu2023nextgpt, sun2023emu, alayrac2022flamingo, zhu2023ghost, li2023videochat, lai2023lisa, yang2023gpt-4v, chen2022pali, li2023otter, zhang2023video-llama, li2023monkey, ye2023mplugdocowl}, which aim to enhance language models with the capability to process and interpret visual information. Flamingo~\cite{alayrac2022flamingo} uses the visual and language inputs as prompts and shows remarkable few-shot performance for visual question answering.
Subsequently, GPT-4~\cite{openai2023gpt4}, LLaVA series~\cite{liu2023llava, lu2023empirical, liu2023improved} and MiniGPT-4 \cite{zhu2023minigpt4} have brought in visual instruction tuning, to improve the instruction-following ability of VLLMs. 
Concurrently, models such as VisionLLM \cite{wang2023visionllm}, KOSMOS-2 \cite{peng2023kosmos2}, and Qwen-VL \etal \cite{bai2023qwenvl, wang2023allseeing, chen2023shikra} have improved VLLMs with visual grounding capabilities, facilitating tasks such as region description and localization. 
Many API-based methods~\cite{2023interngpt,wu2023visual,shen2023hugginggpt,yang2023mmreact,suris2023vipergpt,yang2023gpt4tools,liu2023controlllm} have also attempted to integrate vision APIs with LLMs for solving vision-centric tasks.
Additionally, PaLM-E~\cite{driess2023palme} and EmbodiedGPT~\cite{mu2023embodiedgpt} represent advanced efforts in adapting VLLMs for embodied applications, significantly expanding their potential applications.
These works showcase that VLLMs have achieved significant breakthroughs. 
However, the progress of vision and vision-language foundation models, equally essential for VLLMs, has not kept pace.

\section{Proposed Method}
\label{sec:method}

\subsection{Overall Architecture}

As depicted in Figure \ref{fig:training}, unlike traditional vision-only backbones~\cite{he2016deep, liu2021swin, wang2023internimage} and dual-encoder models~\cite{radford2021clip, openclip, sun2023evaclip}, the proposed \modelname is designed with a vision encoder InternViT-6B and a language middleware QLLaMA.
Specifically, InternViT-6B is a vision transformer with 6 billion parameters, customized to achieve a favorable trade-off between performance and efficiency. 
QLLaMA is a language middleware with 8 billion parameters, initialized with a multilingual-enhanced LLaMA \cite{cui2023chinesellama}. It could provide robust multilingual representation for image-text contrastive learning, or serve as a bridge to connect the vision encoder and the off-the-shelf LLM decoder.

To align the two large-scale components with substantial gaps in modalities and structures,
we introduce a progressive alignment training strategy.
The training strategy is conducted progressively, beginning with contrastive learning on large-scale noisy data, and gradually moving towards generative learning on exquisite and high-quality data.
In this way, we ensure the effective organization and full utilization of web-scale image-text data from a variety of sources.
Then, equipped with the aligned vision encoder and language middleware, our model functions like a Swiss Army knife. 
It boasts a flexible composition that can be adapted for a wide array of generic visual-linguistic tasks. These tasks range from visual perception and image/video-text retrieval to image captioning, visual question answering, and multi-modal dialogue, among others.

\begin{table}[tp]
\scriptsize
\renewcommand\arraystretch{1.0} 
    \centering
    \setlength\tabcolsep{6.1pt}
    \begin{tabular}{l|cccc|c}
        name & width & depth & MLP & \#heads & \#param (M) \\
        \hline

        ViT-G~\cite{zhai2022scaling} & 1664 & 48 & 8192 & 16 & 1843 \\
        
        ViT-e~\cite{chen2022pali} & 1792 & 56 & 15360 & 16 & 3926 \\

        EVA-02-ViT-E~\cite{sun2023evaclip} & 1792 & 64 & 15360 & 16 & 4400 \\

        ViT-6.5B~\cite{singh2023maws} & 4096 & 32 & 16384 & 32 & 6440 \\
        ViT-22B~\cite{dehghani2023vit22b} & 6144 & 48 & 24576 & 48 & 21743 \\

        \rowcolor{gray!15}
        InternViT-6B (ours) & 3200 & 48 & 12800 & 25 & 5903 \\

    \end{tabular}
\caption{\textbf{Architecture details of the InternViT-6B model.}
}
\label{tab:model_config}
\end{table}

\begin{figure*}[t!]
    \centering
    \includegraphics[width=1\textwidth]{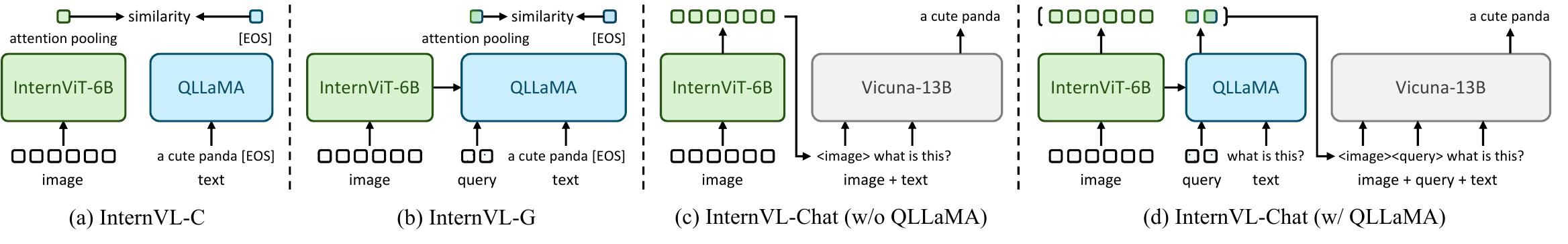}
    \caption{\textbf{Different ways to use InternVL.}
    By flexibly combining the vision encoder and the language middleware, \modelname can support various vision-language tasks, including contrastive tasks, generative tasks, and multi-modal dialogue.
    } 
    \label{fig:inference_mode}
    \vspace{-1em}
\end{figure*}

\subsection{Model Design}
\label{sec:model_design}

\noindent
\textbf{Large-Scale Vision Encoder: InternViT-6B.} 
We implement the vision encoder of \modelname with vanilla vision transformer (ViT)~\cite{dosovitskiy2020image}.
To match the scale of LLMs, we scale up the vision encoder to 6 billion parameters, resulting in the InternViT-6B model.
To obtain a good trade-off between accuracy, speed, and stability, we conduct a hyperparameter search for InternViT-6B. 
We vary the model depth within \{32, 48, 64, 80\}, the head dimension within \{64, 128\}, and the MLP ratio within \{4, 8\}. The model width and the head number are calculated based on the given model scale and other hyperparameters.
%

We employ contrastive learning on a 100M subset of the LAION-en dataset \cite{schuhmann2022laion5b} to measure the accuracy, speed, and stability of InternViT-6B variants with different configurations.
We report the following findings: (1) \emph{Speed.} 
For different model settings, when computation is not saturated, the models with smaller depths exhibit faster speed per image. However, as the GPU computation is fully utilized, the speed difference becomes negligible; 
(2) \emph{Accuracy.} 
With the same number of parameters, the depth, head dimension, and MLP ratio have little impact on the performance. Based on these findings, we identified the most stable configuration for our final model, as shown in Table \ref{tab:model_config}.

\noindent
\textbf{Language Middleware: QLLaMA.} 
The language middleware QLLaMA is proposed to align visual and linguistic features. As shown in Figure \ref{fig:training}, QLLaMA is developed based on the pre-trained multilingual LLaMA~\cite{cui2023chinesellama}, and newly added 96 learnable queries and cross-attention layers (1 billion parameters) that are randomly initialized. 
This manner allows QLLaMA to smoothly integrate visual elements into the language model, thereby enhancing the coherence and effectiveness of the combined features.

Compared to recently popular approaches~\cite{li2023blip2,liu2023llava} that use lightweight ``glue'' layers, such as QFormer~\cite{li2023blip2} and linear layers~\cite{liu2023llava} to connect vision encoder and LLMs,
our method has three advantages: (1) By initializing with the pre-trained weights of \cite{cui2023chinesellama}, QLLaMA can transform image tokens generated by InternViT-6B into the representation that is aligned with the LLMs; (2) QLLaMA has 8 billion parameters for vision-language alignment, which are 42 times larger than the QFormer. Therefore, even with a frozen LLM decoder, \modelname can achieve promising performance on multi-modal dialogue tasks.
(3) It can also be applied to contrastive learning, providing a powerful text representation for image-text alignment tasks, such as zero-shot image classification and image-text retrieval.

\noindent\textbf{``Swiss Army Knife'' Model: \modelname.} By flexibly combining the vision encoder and the language middleware, \modelname can support various vision or vision-language tasks. 

\noindent
(1) \emph{For visual perception tasks}, the vision encoder of \modelname, \ie InternViT-6B, can be used as the backbone for vision tasks. 
Given an input image $I\in\mathbb{R}^{H\times W\times 3}$, our model can generate a feature map $F\in \mathbb{R}^{H/14\times W/14\times D}$ for dense prediction tasks, or work with global average pooling and linear projection to make image classification.

\begin{table}[t]
    \scriptsize
    \centering
    \renewcommand\arraystretch{1.0} 
    \setlength{\tabcolsep}{1.1mm}    
    \begin{tabular}{l|cc|cc|cc}
          & \multicolumn{2}{c|}{characteristics} & \multicolumn{2}{c|}{stage 1}  & \multicolumn{2}{c}{stage 2} \\
         \multirow{-2}{*}{dataset} & language & original & cleaned & remain  & cleaned & remain \\
         \hline
         LAION-en~\cite{schuhmann2022laion5b}  & \multirow{6}{*}{English} & 2.3B & 1.94B & 84.3\% &  91M & 4.0\% \\
         LAION-COCO~\cite{schuhmann2022laioncoco}  &  & 663M     & 550M   & 83.0\% & 550M & 83.0\% \\
         COYO~\cite{byeon2022coyo}        &  & 747M     & 535M   & 71.6\% & 200M & 26.8\% \\
         CC12M~\cite{changpinyo2021cc12m}       &  & 12.4M      & 11.1M  & 89.5\% & 11.1M  & 89.5\% \\
         CC3M~\cite{sharma2018cc3m}        &  & 3.0M       & 2.6M   & 86.7\% & 2.6M   & 86.7\%  \\
         SBU~\cite{ordonez2011sbu}
         &  & 1.0M & 1.0M & 100\% & 1.0M & 100\% \\
         
         \rowcolor{gray!15}
         Wukong~\cite{gu2022wukong}& \multirow{1}{*}{Chinese} & 100M  & 69.4M & 69.4\% & 69.4M & 69.4\% \\
         
         LAION-multi~\cite{schuhmann2022laion5b} & \multirow{1}{*}{Multi} & 2.2B  & 1.87B & 85.0\% & 100M & 4.5\%\\
         \hline
          Total      &  Multi & 6.03B     & 4.98B    & 82.6\% & 1.03B & 17.0\%\\
    \end{tabular}
    \caption{\textbf{Details of the training data for \modelname in stage 1 and stage 2.}
    Among them, LAION-en~\cite{schuhmann2022laion5b}, LAION-multi~\cite{schuhmann2022laion5b}, COYO~\cite{byeon2022coyo}, and Wukong~\cite{gu2022wukong} are web-scale image-text pairs data.
    LAION-COCO~\cite{schuhmann2022laioncoco} is a synthetic dataset with high-quality captions from LAION-en. CC12M~\cite{changpinyo2021cc12m}, CC3M~\cite{sharma2018cc3m}, SBU~\cite{ordonez2011sbu} are academic caption datasets. ``Multi" means multilingual.
    } 
    \label{tab:stage1_data}
\end{table}

\noindent
(2) \emph{For contrastive tasks}, as shown in Figure \ref{fig:inference_mode} (a) (b),
we introduce two inference modes:
\textbf{\modelname-C} and \textbf{\modelname-G}, using the vision encoder or the combination of InternViT and QLLaMA to encode visual features. 
Specifically, we apply attention pooling to the visual features of InternViT or the query features of QLLaMA, to calculate the global visual feature $I_{f}$. 
Besides, we encode text as $T_{f}$ by extracting the feature from the \texttt{[EOS]} token of QLLaMA.
By computing similarity scores between $I_{f}$ and $T_{f}$, we support various contrastive tasks such as image-text retrieval.

\noindent
(3) \emph{For generative tasks}, unlike QFormer~\cite{li2022blip}, QLLaMA inherently has promising image captioning abilities thanks to its scaled-up parameters.
The queries of QLLaMA reorganize the visual representations from InternViT-6B and play as the prefix texts for QLLaMA. The subsequent text tokens are generated one by one sequentially.

\noindent
(4) \emph{For multi-modal dialogue}, we introduce  \textbf{InternVL-Chat}, leveraging \modelname as the visual component to connect with LLMs.
For this purpose, we have two distinct configurations.
One option is to employ the InternViT-6B independently, as shown in Figure \ref{fig:inference_mode} (c).
The alternative is to employ the complete InternVL model concurrently, as illustrated in Figure~\ref{fig:inference_mode} (d).

\begin{table}[t]\scriptsize
    \centering

    \renewcommand{\arraystretch}{1.0}
    \setlength{\tabcolsep}{1.5mm} 
    \begin{tabular}{l|c|l}
         
         task & \#samples & dataset \\
         \hline
         
         Captioning & 588K  & COCO Caption~\cite{chen2015cococaption}, TextCaps~\cite{sidorov2020textcaps}\\

         \rowcolor{gray!15}
            &  & VQAv2~\cite{goyal2017vqav2}, OKVQA~\cite{marino2019okvqa}, A-OKVQA~\cite{schwenk2022aokvqa}, \\
         \rowcolor{gray!15}
         \multirow{-2}{*}{VQA} & \multirow{-2}{*}{1.1M}    & IconQA~\cite{lu2021iconqa}, AI2D~\cite{kembhavi2016ai2d}, GQA~\cite{hudson2019gqa} \\
         
         &  & OCR-VQA~\cite{mishra2019ocrvqa}, ChartQA~\cite{masry2022chartqa}, DocVQA~\cite{clark2017docqa}, \\ 
         & & ST-VQA~\cite{biten2019stvqa}, EST-VQA~\cite{wang2020estvqa}, InfoVQA~\cite{mathew2022infographicvqa}, \\
         \multirow{-3}{*}{OCR} & \multirow{-3}{*}{294K} &LLaVAR~\cite{zhang2023llavar} \\

         \rowcolor{gray!15}
         Grounding & 323K &  RefCOCO/+/g~\cite{yu2016refcoco,mao2016refcocog}, Toloka~\cite{ustalov2023toloka} \\

         Grounded Cap. & 284K  & RefCOCO/+/g~\cite{yu2016refcoco,mao2016refcocog}\\

         \rowcolor{gray!15}
         &  & LLaVA-150K~\cite{liu2023llava}, SVIT~\cite{zhao2023svit}, VisDial~\cite{das2017visdial}, \\
         \rowcolor{gray!15}
         \multirow{-2}{*}{Conversation} & \multirow{-2}{*}{1.4M} & LRV-Instruction~\cite{liu2023lrv-instruction}, LLaVA-Mix-665K~\cite{liu2023improved} \\
    \end{tabular}
    \caption{\textbf{Details of the training data for \modelname in stage 3. } 
    We collect a wide range of high-quality instruction data, totaling approximately 4 million samples. For a fair comparison, we only use the training split of these datasets.
    }
    \label{tab:stage3_data}
\end{table} 

\subsection{Alignment Strategy}

As shown in Figure~\ref{fig:training}, the training of \modelname consists of three progressive stages, including vision-language contrastive training, vision-language generative training, and supervised fine-tuning. These stages effectively leverage public data from diverse sources, ranging from noisy image-text pairs on the web to high-quality caption, VQA, and multi-modal dialogue datasets.

\noindent
\textbf{Vision-Language Contrastive Training.} 
In the first stage, we conduct contrastive learning to align InternViT-6B with a multilingual LLaMA-7B \cite{cui2023chinesellama} on web-scale, noisy image-text pairs. 
The data are all publicly available and comprise multilingual content, including LAION-en \cite{schuhmann2022laion5b}, LAION-multi \cite{schuhmann2022laion5b}, LAION-COCO \cite{schuhmann2022laioncoco}, COYO~\cite{byeon2022coyo}, Wukong \cite{gu2022wukong}, etc. 
We use the combination of these datasets and filter out some extremely low-quality data to train our model. 
As summarized in Table~\ref{tab:stage1_data}, the original dataset contains 6.03 billion image-text pairs, and 4.98 billion remains after cleaning. 
More details about data preparation will be provided in the supplementary materials.

During training, we adopt the LLaMA-7B to encode the text as $T_{f}$, and use InternViT-6B to extract the visual feature $I_{f}$. Following the objective function of CLIP \cite{radford2021clip}, we minimize a symmetric cross-entropy loss on the similarity scores of image-text pairs in a batch. This stage allows \modelname to excel on contrastive tasks like zero-shot image classification and image-text retrieval, and the vision encoder of this stage can also perform well on visual perception tasks like semantic segmentation.

\noindent
\textbf{Vision-Language Generative Training}. 
In the second stage of training, we connect InternViT-6B with QLLaMA and adopt a generative training strategy. 
Specifically, QLLaMA inherits the weights of LLaMA-7B in the first stage.
We keep both InternViT-6B and QLLaMA frozen and only train the newly added learnable queries and cross-attention layers with filtered, high-quality data.
Table~\ref{tab:stage1_data} summarizes the datasets for the second stage.
It can be seen that we further filtered out data with low-quality captions, reducing it from 4.98 billion in the first stage to 1.03 billion.

Following the loss function of BLIP-2~\cite{li2023blip2}, the loss in this stage is computed as the sum of three components: image-text contrastive (ITC) loss, image-text matching (ITM) loss, and image-grounded text generation (ITG) loss. 
This enables the queries to extract powerful visual representations, and further align feature space with LLMs, attributable to the effective training objectives and the utilization of our large-scale, LLM-initialized QLLaMA.

\begin{table}[t]\scriptsize
\renewcommand{\arraystretch}{1.0}
\centering

\setlength\tabcolsep{1.4pt}
\begin{tabular}{l|c|cccccc|c}
    method & \#param & IN-1K & IN-ReaL & IN-V2 & IN-A & IN-R & IN-Ske & avg. \\
    
    \hline
    
    OpenCLIP-H~\cite{openclip} & 0.6B &
    84.4 & 88.4 & 75.5 & $-$ & $-$ & $-$ & $-$ \\
    
    OpenCLIP-G~\cite{openclip} & 1.8B  &
    86.2 & 89.4 & 77.2 & 63.8 & 87.8 & 66.4 & 78.5 \\

    DINOv2-g~\cite{oquab2023dinov2} & 1.1B  &
    86.5 & 89.6 & 78.4 & 75.9 & 78.8 & 62.5 & 78.6 \\

    EVA-01-CLIP-g~\cite{fang2022eva} & 1.1B  &
    86.5 & 89.3 & 77.4 & 70.5 & 87.7 & 63.1 & 79.1 \\

    MAWS-ViT-6.5B~\cite{singh2023maws} & 6.5B & 
    87.8 & --  & -- & -- & -- & -- & -- \\

    \textcolor{gray}{ViT-22B$^*$~\cite{dehghani2023vit22b}}  & \textcolor{gray}{21.7B} &
    \textcolor{gray}{89.5} & \textcolor{gray}{90.9} & \textcolor{gray}{83.2} & \textcolor{gray}{83.8} & \textcolor{gray}{87.4} & \textcolor{gray}{$-$} & $-$ \\
    
    \rowcolor{gray!15}
    InternViT-6B (ours) & 5.9B &
    \textbf{88.2} & \textbf{90.4} &  \textbf{79.9}  & \textbf{77.5} &  \textbf{89.8}  &  \textbf{69.1} & \textbf{82.5} \\

\end{tabular}
\caption{\textbf{Linear evaluation on image classification.}~We report the top-1 accuracy on ImageNet-1K \cite{deng2009imagenet} and its variants \cite{beyer2020imagenetreal,recht2019imagenetv2,hendrycks2021imagenet_a,hendrycks2021imagenet_r,wang2019imagenet_sketch}.
$^*$ViT-22B \cite{dehghani2023vit22b} uses the private JFT-3B dataset~\cite{zhai2022scaling}.
}
\label{tab:img_cls}
\end{table}

\begin{table}[t]\scriptsize
\renewcommand{\arraystretch}{1.0}

\begin{subtable}{0.47\textwidth}
    \centering

    \setlength\tabcolsep{4.3pt}
    \begin{tabular}{l|c|c|ccccc}
        method & \#param & crop size & $1/16$ & $1/8$ & $1/4$ & $1/2$ & $1$ \\
        \hline
        ViT-L~\cite{touvron2022deit3} & 
        0.3B & 504$^2$ & 36.1 & 41.3 & 45.6 & 48.4 & 51.9 \\
        
        ViT-G~\cite{zhai2022scaling} & 
        1.8B & 504$^2$ & 42.4 & 47.0 & 50.2 & 52.4 & 55.6 \\
        
        ViT-22B~\cite{dehghani2023vit22b} & 
        21.7B & 504$^2$ & 44.7 & 47.2 & 50.6 & 52.5 & 54.9 \\
        
        \rowcolor{gray!15}
        InternViT-6B (ours) & 
        5.9B & 504$^2$ & \textbf{46.5} & \textbf{50.0} & \textbf{53.3} & \textbf{55.8} & \textbf{57.2} \\
         
    \end{tabular}
    \caption{Few-shot semantic segmentation with limited training data.
Following ViT-22B~\cite{dehghani2023vit22b}, we fine-tune the InternViT-6B with a linear classifier.
}
\label{tab:few_shot_seg}
\end{subtable}
\begin{subtable}{0.47\textwidth}
    \setlength\tabcolsep{4.5pt}
    \begin{tabular}{l|c|c|c|cc}
        method & decoder & \#param (train/total) & crop size & mIoU \\ 
        
        \hline
        OpenCLIP-G\textsubscript{frozen}~\cite{openclip} & Linear & 0.3M / 1.8B & 512$^2$ & 39.3 \\ 

        
        ViT-22B\textsubscript{frozen}~\cite{dehghani2023vit22b} & Linear & 0.9M / 21.7B & 504$^2$ & 34.6 \\

        \rowcolor{gray!15}
        InternViT-6B\textsubscript{frozen} (ours) & Linear & 0.5M / 5.9B & 504$^2$ & \textbf{47.2}  \\

        \hline
        ViT-22B\textsubscript{frozen}~\cite{dehghani2023vit22b} & UperNet & 
        0.8B / 22.5B & 504$^2$ & 52.7 \\

        \rowcolor{gray!15}
        InternViT-6B\textsubscript{frozen} (ours) & UperNet& 
        0.4B / 6.3B & 504$^2$ & \textbf{54.9} \\

        \hline
    
        ViT-22B~\cite{dehghani2023vit22b} & UperNet & 
        22.5B / 22.5B & 504$^2$ & 55.3 \\

        \rowcolor{gray!15}
        InternViT-6B (ours) & UperNet & 
        6.3B / 6.3B & 504$^2$ & \textbf{58.9} \\

    \end{tabular}
    \centering
        \caption{Semantic segmentation performance in three different settings, from top to bottom: linear probing, head tuning, and full-parameter tuning.
    }
    \label{tab:linear_seg}
\end{subtable}
\caption{\textbf{Semantic segmentation on ADE20K.} 
Results show that InternViT-6B has better pixel-level perceptual capacity.
}
\label{tab:sem_seg}
\end{table}

\noindent
\textbf{Supervised Fine-tuning.}
To demonstrate the benefits of \modelname in creating multi-modal dialogue systems, we connect it with an off-the-shelf LLM decoder (\eg, Vicuna~\cite{zheng2023vicuna} or InternLM~\cite{2023internlm}) through an MLP layer, and conduct supervised fine-tuning (SFT).
As detailed in Table \ref{tab:stage3_data}, we collect a wide range of high-quality instruction data, totaling approximately 4 million samples.
For non-dialogue datasets, we follow the method described in \cite{liu2023improved} for conversion.
Owing to the similar feature space of QLLaMA and LLMs, we can achieve robust performance even when freezing the LLM decoder, choosing to train just the MLP layer or both the MLP layer and QLLaMA. 
This approach not only expedites the SFT process but also maintains the original language capabilities of the LLMs.

\begin{table*}[!h]\scriptsize
    \centering

    \renewcommand{\arraystretch}{1.0}
    
    \begin{subtable}{0.53\linewidth}
        \centering
        \setlength\tabcolsep{3pt}
        \begin{tabular}{l|cccccc|c|c}
        method &  IN-1K &  IN-A &  IN-R &  IN-V2 &  IN-Sketch &  ObjectNet & $\Delta${$\downarrow$} & avg. \\
        \hline

         OpenCLIP-H~\cite{openclip} &
         78.0 &  59.3 &  89.3 &  70.9 &  66.6 &  69.7 & 5.7 &  72.3\\
         
         OpenCLIP-g~\cite{openclip} & 
         78.5 &  60.8 &  90.2 &  71.7 &  67.5 &  69.2 & 5.5 &  73.0 \\
         
         OpenAI CLIP-L+~\cite{radford2021clip} &
         76.6 &  77.5 &  89.0 &  70.9 &  61.0 &  72.0 & 2.1 &  74.5 \\
         
         EVA-01-CLIP-g~\cite{sun2023evaclip} &
         78.5 &  73.6 &  92.5 &  71.5 &  67.3 &  72.3 & 2.5 &  76.0 \\
         
         OpenCLIP-G~\cite{openclip} & 
         80.1 &  69.3 &  92.1 &  73.6 &  68.9 &  73.0 & 3.9 &  76.2\\
         
         EVA-01-CLIP-g+~\cite{sun2023evaclip} & 
         79.3 &  74.1 &  92.5 &  72.1 &  68.1 &  75.3 & 2.4 &  76.9\\
         
         
         
         MAWS-ViT-2B~\cite{singh2023maws} & 
         81.9 & -- & -- & -- & -- & -- & -- & -- \\
         
         EVA-02-CLIP-E+~\cite{sun2023evaclip} & 
         82.0 &  82.1 &  94.5 &  75.7 &  71.6 &  79.6 &  1.1 &  80.9\\

         \textcolor{gray}{CoCa$^*$~\cite{yu2022coca}} & 
         \textcolor{gray}{86.3} &  \textcolor{gray}{90.2} &  \textcolor{gray}{96.5} &  \textcolor{gray}{80.7} &  \textcolor{gray}{77.6} &  \textcolor{gray}{82.7} & \textcolor{gray}{0.6}  &  \textcolor{gray}{85.7} \\

         \textcolor{gray}{LiT-22B$^*$~\cite{dehghani2023vit22b,zhai2022lit}} & 
         \textcolor{gray}{85.9} &  \textcolor{gray}{90.1} &  \textcolor{gray}{96.0} &  \textcolor{gray}{80.9} &  \textcolor{gray}{$-$} &  \textcolor{gray}{87.6} & \textcolor{gray}{$-$}  &  \textcolor{gray}{$-$} \\
        
         \rowcolor{gray!15}
         InternVL-C (ours) &
         \textbf{83.2} &	\textbf{83.8} &	\textbf{95.5} &	\textbf{77.3} &	\textbf{73.9} &	\textbf{80.6} &	\textbf{0.8} &  \textbf{82.4} \\
    \end{tabular}
    \caption{ImageNet variants~\cite{deng2009imagenet,hendrycks2021imagenet_a,hendrycks2021imagenet_r,recht2019imagenetv2,wang2019imagenet_sketch} and ObjectNet~\cite{barbu2019objectnet}.}
    \label{subtable:in1k_variant}
    \end{subtable}%
    \hspace{2.5em}
    \begin{subtable}{0.42\linewidth}  
        \centering
        \setlength\tabcolsep{4pt}
      \begin{tabular}{l|ccccc|c}
        method &  EN &  ZH &  JP &  AR &  IT &  avg. \\

        \hline
        M-CLIP~\cite{carlsson2022mclip}  &
        $-$	& $-$ & $-$ & $-$ & 20.2 & $-$ \\
        
        CLIP-Italian~\cite{bianchi2021clip_italian}  &
        $-$	& $-$ & $-$ & $-$ & 22.1 & $-$ \\

        Japanese-CLIP-ViT-B~\cite{japanese-clip} &
        $-$	& $-$ & 54.6 & $-$ & $-$ & $-$ \\
        
        Taiyi-CLIP-ViT-H~\cite{fengshenbang} &
        $-$	& 54.4 & $-$ & $-$ & $-$ & $-$ \\

        WuKong-ViT-L-G~\cite{gu2022wukong} &
        $-$	& 57.5 & $-$ & $-$ & $-$ & $-$ \\

        
        CN-CLIP-ViT-H~\cite{yang2022cnclip} &
        $-$	& 59.6 & $-$ & $-$ & $-$ & $-$ \\

        AltCLIP-ViT-L~\cite{chen2022altclip} &
        74.5 & 59.6 & $-$ & $-$ & $-$ & $-$ \\

        EVA-02-CLIP-E+~\cite{sun2023evaclip} & 
        82.0 & 3.6 & 5.0 & 0.2 & 41.2 & $-$ \\

        
        OpenCLIP-XLM-R-B~\cite{openclip} &
        62.3 &	42.7 &	37.9 &	26.5 &	43.7 &	42.6  \\

        OpenCLIP-XLM-R-H~\cite{openclip} &
        77.0 &	55.7 &	53.1 &	37.0 &	56.8 &	55.9  \\
        
        \rowcolor{gray!15}
        InternVL-C (ours) &
        \textbf{83.2} &	\textbf{64.5} &	\textbf{61.5} &	\textbf{44.9} &	\textbf{65.7} &	\textbf{64.0}  \\
    \end{tabular}
    \caption{Multilingual ImageNet-1K~\cite{deng2009imagenet,laion_ai_2023_clip}.}
    \label{subtable:multilingual_in1k}
    \end{subtable}%
    \caption{\textbf{Comparison of zero-shot image classification performance.} 
    ``{$\Delta$\scriptsize$\downarrow$}'': The gap between the averaged top-1 accuracy and the IN-1K top-1 accuracy. 
    $^*$CoCa~\cite{yu2022coca} and LiT-22B~\cite{dehghani2023vit22b} use the private JFT-3B dataset~\cite{zhai2022scaling} during training.
    Multilingual evaluation involves 5 languages, including English (EN), Chinese (ZH), Japanese (JP), Arabic (AR), and Italian (IT).
    }
    \label{table:zs_in1k}
\end{table*}

\begin{table*}[t]\scriptsize
\renewcommand{\arraystretch}{1.0}
\centering

\setlength\tabcolsep{6.5pt}
\begin{tabular}{l|c|ccc|ccc|ccc|ccc|c}
    &  & \multicolumn{6}{c|}{Flickr30K (English, 1K test set)~\cite{plummer2015flickr30k}} & \multicolumn{6}{c|}{COCO (English, 5K test set)~\cite{chen2015cococaption}} & \\
    & multi- & \multicolumn{3}{c|}{Image $\rightarrow$ Text} & \multicolumn{3}{c|}{Text $\rightarrow$ Image} & \multicolumn{3}{c|}{Image $\rightarrow$ Text} & \multicolumn{3}{c|}{Text $\rightarrow$ Image} & \\
    \multirow{-2}{*}{method} & lingual & R@1 &  R@5 &  R@10 &  R@1 &  R@5 &  R@10 &  R@1 &  R@5 &  R@10 &  R@1 &  R@5 &  R@10 & \multirow{-2}{*}{avg.} \\
    \hline
     

     Florence~\citep{yuan2021florence} & $\times$ &
     90.9 & 99.1 & $-$ & 76.7 & 93.6 & $-$ & 64.7 & 85.9 & $-$ & 47.2 & 71.4 & $-$ & $-$ \\

     ONE-PEACE~\citep{wang2023onepeace} & $\times$ &
     90.9 & 98.8 & 99.8 & 77.2 & 93.5 & 96.2 & 64.7 & 86.0 & 91.9 & 48.0 & 71.5 & 79.6 & 83.2 \\   
     
     OpenCLIP-H~\citep{openclip} & $\times$ &
     90.8 & 99.3 & 99.7 & 77.8 & 94.1 & 96.6 & 66.0 & 86.1 & 91.9 & 49.5 & 73.4 & 81.5 & 83.9 \\
     
     OpenCLIP-g~\citep{openclip} & $\times$ &
     91.4 & 99.2 & 99.6 & 77.7 & 94.1 & 96.9 & 66.4 & 86.0 & 91.8 & 48.8 & 73.3 & 81.5 & 83.9 \\

     OpenCLIP-XLM-R-H~\cite{openclip}  & $\checkmark$ &
     91.8 & 99.4 & 99.8 & 77.8 & 94.1 & 96.5 & 65.9 & 86.2 & 92.2 & 49.3 & 73.2 & 81.5 & 84.0 \\ 

     EVA-01-CLIP-g+~\citep{sun2023evaclip} & $\times$ & 
     91.6 & 99.3 & 99.8 & 78.9 & 94.5 & 96.9 & 68.2 & 87.5 & 92.5 & 50.3 & 74.0 & 82.1 & 84.6 \\
    
     CoCa~\citep{yu2022coca} & $\times$ &
     92.5 & 99.5 & 99.9 & 80.4 & 95.7 & 97.7 & 66.3 & 86.2 & 91.8 & 51.2 & 74.2 & 82.0 & 84.8 \\ 
     
     OpenCLIP-G~\citep{openclip} & $\times$ &
     92.9 & 99.3 & 99.8 & 79.5 & 95.0 & 97.1 & 67.3 & 86.9 & 92.6 & 51.4 & 74.9 & 83.0 & 85.0 \\

     
     
     EVA-02-CLIP-E+~\citep{sun2023evaclip} & $\times$ &
     93.9 & 99.4 & 99.8 & 78.8 & 94.2 & 96.8 & 68.8 & 87.8 & 92.8 & 51.1 & 75.0 & 82.7 & 85.1 \\

     \textcolor{gray}{BLIP-2$^\dagger$~\cite{li2023blip2}} & \textcolor{gray}{$\times$} & \textcolor{gray}{97.6} & \textcolor{gray}{100.0} & \textcolor{gray}{100.0} & \textcolor{gray}{89.7} & \textcolor{gray}{98.1} & \textcolor{gray}{98.9} & \textcolor{gray}{$-$} & \textcolor{gray}{$-$} & \textcolor{gray}{$-$} & \textcolor{gray}{$-$} & \textcolor{gray}{$-$} & \textcolor{gray}{$-$} & \textcolor{gray}{$-$} \\
     
     
     \rowcolor{gray!15}
     InternVL-C (ours) & $\checkmark$ &
     94.7 & 99.6 & 99.9 & 81.7 & 96.0 & 98.2 & 70.6 & 89.0 & 93.5 & 54.1 & 77.3 & 84.6 & 86.6 \\  
     \rowcolor{gray!15}
     InternVL-G (ours)  & $\checkmark$ &
     \textbf{95.7} & \textbf{99.7} & \textbf{99.9} & \textbf{85.0} & \textbf{97.0} & \textbf{98.6} & \textbf{74.9} & \textbf{91.3} & \textbf{95.2} & \textbf{58.6} & \textbf{81.3} & \textbf{88.0} & \textbf{88.8} \\
     
     \multicolumn{15}{c}{ }\\
     \multicolumn{1}{l|}{method} & \multicolumn{1}{c|}{} & \multicolumn{6}{c|}{ Flickr30K-CN (Chinese, 1K test set)~\citep{lan2017flickrcn}} & \multicolumn{6}{c|}{COCO-CN (Chinese, 1K test set)~\citep{li2019cococn}} & avg.\\

     \hline
     
     WuKong-ViT-L~\citep{gu2022wukong} & $\times$ &
     76.1 & 94.8 & 97.5 & 51.7 & 78.9 & 86.3 & 55.2 & 81.0 & 90.6 & 53.4 & 80.2 & 90.1 & 78.0 \\
     
     R2D2-ViT-L~\citep{xie2022zero}  & $\times$ &
     77.6 & 96.7 & 98.9 & 60.9 & 86.8 & 92.7 & 63.3 & 89.3 & 95.7 & 56.4 & 85.0 & 93.1 & 83.0 \\
     
     Taiyi-CLIP-ViT-H~\citep{fengshenbang}  & $\times$ &
      $-$ & $-$  & $-$  &  $-$ &  $-$ &  $-$ &  $-$ &  $-$ &  $-$ & 60.0 & 84.0 &	93.3 & $-$ \\


     AltCLIP-ViT-H~\citep{chen2022altclip}  & $\checkmark$ &
     88.9 & 98.5 & 99.5 & 74.5 & 92.0 & 95.5 & $-$ & $-$  & $-$  &  $-$ &  $-$ &  $-$ &  $-$ \\
     
     CN-CLIP-ViT-H~\citep{yang2022cnclip}  & $\times$ &
     81.6 & 97.5 & 98.8 & 71.2 & 91.4 & 95.5 & 63.0 & 86.6 & 92.9 & 69.2 & 89.9 & 96.1 & 86.1 \\ 
     
     OpenCLIP-XLM-R-H~\cite{openclip}  & $\checkmark$ &
     86.1 & 97.5 & 99.2 & 71.0 & 90.5 & 94.9 & 70.0 & 91.5 & 97.0 & 66.1 & 90.8 & 96.0 & 87.6 \\  

             
     \rowcolor{gray!15}
     InternVL-C (ours)  & $\checkmark$ &
     90.3 & 98.8 & 99.7 & 75.1 & 92.9 & 96.4 & 68.8 & 92.0 & 96.7 & 68.9 & 91.9 & 96.5 & 89.0 \\  
     
     \rowcolor{gray!15}
     InternVL-G (ours) & $\checkmark$ &
     \textbf{92.9} & \textbf{99.4} & \textbf{99.8} & \textbf{77.7} & \textbf{94.8} & \textbf{97.3} & \textbf{71.4} & \textbf{93.9} & \textbf{97.7} & \textbf{73.8} & \textbf{94.4} & \textbf{98.1} & \textbf{90.9} \\  

\end{tabular}
\caption{\textbf{Comparison of zero-shot image-text retrieval performance.}
We evaluate the retrieval capability in English using the Flickr30K~\cite{plummer2015flickr30k} and COCO~\cite{chen2015cococaption}, as well as in Chinese using Flickr30K-CN~\cite{lan2017flickrcn} and COCO-CN~\cite{li2019cococn}.
$^\dagger$BLIP-2 \cite{li2023blip2} is finetuned on COCO and zero-shot transferred to Flickr30K, contributing to the enhanced zero-shot performance on Flickr30K.
}
\label{tab: clip zs retrieval}
\end{table*}

\section{Experiments}
\label{sec:experiments}

\subsection{Implementation Details}

\noindent
\textbf{Stage 1.} 
In this stage, the image encoder InternViT-6B is randomly initialized~\cite{bao2021beit}, and the text encoder LLaMA-7B is initialized with the pre-trained weights from \cite{cui2023chinesellama}.
All parameters are fully trainable.

\noindent
\textbf{Stage 2.}
In this stage, InternViT-6B and QLLaMA inherit their weights from the first stage, while the new learnable queries and cross-attention layers in QLLaMA are randomly initialized.
Benefiting from the powerful representations learned in the first stage, we keep both InternViT-6B and QLLaMA frozen and only train the new parameters.

\noindent
\textbf{Stage 3.}
At this stage, we have two different configurations.
One is to use InternViT-6B separately, as shown in Figure~\ref{fig:inference_mode}~(c). 
The other is to use the entire InternVL model simultaneously, as shown in Figure~\ref{fig:inference_mode} (d). 
More details will be provided in the supplementary materials.

\subsection{Visual Perception Benchmarks}

First of all, we validate the visual perception capabilities of InternViT-6B, the most core component of \modelname.

\noindent\textbf{Transfer to Image Classification.}
We evaluate the quality of visual representation produced by InternViT-6B using the ImageNet-1K \cite{deng2009imagenet} dataset. 
Following common practices \cite{he2022mae,oquab2023dinov2,dehghani2023vit22b}, we adopt the linear probing evaluation, \ie training a linear classifier while keeping the backbone frozen. 
In addition to the ImageNet-1K validation set, we also report performance metrics on several ImageNet variants~\cite{beyer2020imagenetreal, recht2019imagenetv2, hendrycks2021imagenet_a, hendrycks2021imagenet_r, wang2019imagenet_sketch}, to benchmark the domain generalization capability.
As shown in Table \ref{tab:img_cls}, InternViT-6B achieves a very significant improvement over previous state-of-the-art methods \cite{fang2022eva,oquab2023dinov2,openclip} on linear probing. 
To our knowledge, this represents the currently best linear evaluation results without the JFT dataset \cite{zhai2022scaling}.

\noindent\textbf{Transfer to Semantic Segmentation.}
To investigate the pixel-level perceptual capacity of InternViT-6B, we conduct extensive experiments of semantic segmentation on the ADE20K~\cite{zhou2017ade20k} dataset.
Following ViT-22B \cite{dehghani2023vit22b}, we begin with few-shot learning experiments, \ie fine-tuning the backbone with a linear head on a limited dataset.
As indicated in Table~\ref{tab:few_shot_seg}, InternViT-6B consistently outperforms ViT-22B across five experiments with varying proportions of training data.
Additionally, Table~\ref{tab:linear_seg} presents our further verification in three distinct settings, including linear probing, head tuning~\cite{xiao2018upernet}, and full-parameter tuning.
Notably, in the case of linear probing, InternViT-6B attains 47.2 mIoU, a substantial +12.6 mIoU improvement over ViT-22B. 
These results underscore the strong out-of-the-box pixel-level perceptual capacity of our InternViT-6B.

\begin{table}[t]\scriptsize
\renewcommand{\arraystretch}{1.0}
    \centering

    \setlength\tabcolsep{4.8pt}
    \begin{tabular}{l|c|cc|cc|cc}
         & & \multicolumn{2}{c|}{K400~\cite{carreira2017k400}} & \multicolumn{2}{c|}{K600~\cite{carreira2018k600}} & \multicolumn{2}{c}{K700~\cite{carreira2019k700}} \\
        \multirow{-2}{*}{method} & \multirow{-2}{*}{\#F} & top-1 & avg. & top-1 & avg. & top-1 & avg. \\
        \hline
        
        OpenCLIP-g~\cite{openclip} &1&
        $-$  & 63.9 & $-$  & 64.1 & $-$  & 56.9 \\
        
        OpenCLIP-G~\cite{openclip} &1&
        $-$  & 65.9 & $-$  & 66.1 & $-$  & 59.2 \\
        
        EVA-01-CLIP-g+~\cite{sun2023evaclip} &1& 
        $-$  & 66.7 &  $-$ & 67.0 & $-$  & 60.9 \\
        
        EVA-02-CLIP-E+~\cite{sun2023evaclip} &1&
        $-$  & 69.8 &  $-$ & 69.3 & $-$  & 63.4 \\

        \rowcolor{gray!15}
        InternVL-C (ours) &1& 
        \textbf{65.9} & \textbf{76.1} & \textbf{65.5} & \textbf{75.5} & \textbf{56.8} & \textbf{67.5} \\
        
        \hline
        
        ViCLIP~\cite{wang2023internvid} & 8&
        64.8 & 75.7 & 62.2 & 73.5 & 54.3 & 66.4 \\

        \rowcolor{gray!15}
        InternVL-C (ours) &8& 
        \textbf{69.1} & \textbf{79.4} & \textbf{68.9} & \textbf{78.8} & \textbf{60.6} & \textbf{71.5} \\

    \end{tabular}

\caption{
\textbf{Comparison of zero-shot video classification results on Kinetics 400/600/700.} 
We report the top-1 accuracy and the mean of top-1 and top-5 accuracy. ``\#F" denotes the number of frames.
}
\label{tab: zs_video}
\end{table}

\begin{table*}[t!]
\scriptsize
\centering

\setlength\tabcolsep{2.6pt}
\renewcommand{\arraystretch}{1.0}
\begin{tabular}{l|lll|cccc|ccc|cccc|cc}
 & visual & glue & &  &  &  & train. & \multicolumn{3}{c|}{image captioning} & \multicolumn{4}{c|}{visual question answering}& \multicolumn{2}{c}{dialogue}\\

\multirow{-2}{*}{method} & encoder & layer & \multirow{-2}{*}{LLM} & \multirow{-2}{*}{Res.} & \multirow{-2}{*}{PT} & \multirow{-2}{*}{SFT} & param & COCO & Flickr & NoCaps & VQA$^\text{v2}$ & GQA & VizWiz  & VQA$^\text{T}$ & MME & POPE \\
\hline

InstructBLIP~\cite{instructblip} &
EVA-g & QFormer & Vicuna-7B & 224 & 129M & 1.2M & 188M &
-- & 82.4 & 123.1 & 
-- & 49.2 & 34.5 & 50.1 &
-- & --  \\

BLIP-2~\cite{li2023blip2} & 
EVA-g & QFormer & Vicuna-13B & 224 & 129M & -- & 188M &
-- & 71.6 & 103.9 &
41.0 & 41.0 & 19.6 & 42.5 &
1293.8 & 85.3 \\

InstructBLIP~\cite{instructblip} & 
EVA-g & QFormer & Vicuna-13B & 224 & 129M & 1.2M & 188M & 
-- & 82.8 & 121.9 & 
-- & 49.5 & 33.4 & 50.7 & 
1212.8 & 78.9 \\

\rowcolor{gray!15}
InternVL-Chat (ours) & 
IViT-6B & QLLaMA & Vicuna-7B & 224 & 1.0B & 4.0M & 64M &
~141.4$^*$ & 89.7 & 120.5 & 
~72.3$^*$ & ~57.7$^*$ & 44.5 & 42.1 &
1298.5 & 85.2  \\

\rowcolor{gray!15}
InternVL-Chat (ours) & 
IViT-6B & QLLaMA & Vicuna-13B & 224 & 1.0B & 4.0M & 90M & 
~142.4$^*$ & 89.9 & 123.1 & 
~71.7$^*$ & ~59.5$^*$ & 54.0 & 49.1 &
1317.2 & 85.4  \\

\hline
Shikra~\cite{chen2023shikra} & 
CLIP-L & Linear & Vicuna-13B &
224 & 600K & 5.5M & 7B & 
~117.5$^*$ & 73.9 & -- &
~77.4$^*$ & -- & -- & -- &
-- & --   \\

IDEFICS-80B~\cite{idefics2023} & 
CLIP-H & Cross-Attn & LLaMA-65B &
224 & 1.6B & -- & 15B &
~91.8$^*$ & 53.7 & 65.0 &
60.0 & 45.2 & 36.0 & 30.9 &
-- & --  \\

IDEFICS-80B-I~\cite{idefics2023} &
CLIP-H & Cross-Attn & LLaMA-65B & 
224 & 353M & 6.7M & 15B &
~117.2$^*$ & 65.3 & 104.5 & 
37.4 & -- & 26.0 & -- & 
 -- & --  \\

Qwen-VL~\cite{bai2023qwenvl} &
CLIP-G & VL-Adapter & Qwen-7B & 
448 & 1.4B$^\dagger$ & 50M$^\dagger$ & 9.6B &
-- & 85.8 & 121.4 & 
~78.8$^*$ & ~59.3$^*$ & 35.2 & 63.8 &
-- & --  \\

Qwen-VL-Chat~\cite{bai2023qwenvl} & 
CLIP-G & VL-Adapter & Qwen-7B &
448 & 1.4B$^\dagger$ & 50M$^\dagger$ & 9.6B & 
-- & 81.0 & 120.2 & 
~78.2$^*$ & ~57.5$^*$ & 38.9 & \textbf{61.5} &
1487.5 & --  \\

LLaVA-1.5~\cite{liu2023improved} & 
CLIP-L$_{336}$ & MLP & Vicuna-7B & 
336 & 558K & 665K & 7B & 
-- & -- & -- &
~78.5$^*$ & ~62.0$^*$ & 50.0 & 58.2 &
1510.7 & 85.9  \\

LLaVA-1.5~\cite{liu2023improved} &
CLIP-L$_{336}$ & MLP & Vicuna-13B &
336 & 558K & 665K & 13B & 
-- & -- & -- &
~80.0$^*$ & ~63.3$^*$ & 53.6 & 61.3 &
1531.3 & 85.9  \\

\rowcolor{gray!15}
InternVL-Chat (ours) &
IViT-6B & MLP & Vicuna-7B &
336 & 558K & 665K & 7B &
-- & -- & -- &
~79.3$^*$ & ~62.9$^*$ & 52.5 & 57.0 & 
1525.1 & 86.4  \\

\rowcolor{gray!15}
InternVL-Chat (ours) &
IViT-6B & MLP & Vicuna-13B & 
336 & 558K & 665K & 13B & 
-- & -- & -- &
~80.2$^*$ & ~63.9$^*$ & 54.6 & 58.7 & 
1546.9 & 87.1  \\

\hline

\rowcolor{gray!15}
InternVL-Chat (ours) &
IViT-6B & QLLaMA & Vicuna-13B & 
336 & 1.0B & 4.0M & 13B & 
~\textbf{146.2}$^*$ & \textbf{92.2} & \textbf{126.2} &
~\textbf{81.2}$^*$ & ~\textbf{66.6}$^*$ & \textbf{58.5} & \textbf{61.5} & 
\textbf{1586.4} & \textbf{87.6}  \\

\end{tabular}
\caption{\textbf{Comparison with SoTA methods on 9 benchmarks.}
Image captioning datasets include: COCO Karpathy test \cite{chen2015cococaption}, Flickr30K Karpathy test \cite{plummer2015flickr30k}, NoCaps val \cite{agrawal2019nocaps}.
VQA datasets include: VQAv2 test-dev \cite{goyal2017vqav2}, GQA test-balanced \cite{hudson2019gqa}, VizWiz test-dev \cite{gurari2018vizwiz}, and TextVQA val \cite{singh2019textvqa}.
$^*$The training annotations of the datasets are observed during training. ``IViT-6B'' represents our InternViT-6B. 
}
\label{tab:sota_results}
\end{table*}

\begin{table}[t]\scriptsize
\renewcommand{\arraystretch}{1.0}
\centering

    \setlength\tabcolsep{2.7pt}
    \begin{tabular}{l|ll|ccc}
        method & glue layer & LLM decoder & COCO & Flickr30K & NoCaps  \\
        \hline

        Flamingo-9B~\cite{alayrac2022flamingo} & 
        Cross-Attn & Chinchilla-7B & 79.4 & 61.5 & -- \\

        Flamingo-80B~\cite{alayrac2022flamingo} & 
        Cross-Attn & Chinchilla-70B & 84.3 & 67.2 & -- \\
        
        KOSMOS-2~\cite{peng2023kosmos2} & 
        Linear & KOSMOS-1 & --  & 66.7 &  -- \\

        PaLI-X-55B~\cite{chen2023palix} & 
        Linear & UL2-32B & -- & -- & \textbf{126.3} \\
        
        BLIP-2~\cite{li2023blip2} & 
        QFormer & Vicuna-13B & -- & 71.6 & 103.9  \\

        InstructBLIP~\cite{instructblip} & 
        QFormer & Vicuna-13B & -- & 82.8 & 121.9 \\
        
        Shikra-13B~\cite{chen2023shikra} &
        Linear & Vicuna-13B & --  & 73.9 &  --  \\
        
        ASM~\cite{wang2023allseeing} &
        QFormer & Husky-7B & --  & \textbf{87.7} & 117.2 \\

        Qwen-VL~\cite{bai2023qwenvl} & 
        VL-Adapter & Qwen-7B & -- & 85.8 & 121.4 \\
        
        Qwen-VL-Chat~\cite{bai2023qwenvl} &
        VL-Adapter & Qwen-7B & --  & 81.0 & 120.2  \\

        Emu~\cite{sun2023emu} & 
        QFormer & LLaMA-13B & 112.4 & -- &  --  \\
        
        Emu-I~\cite{sun2023emu} & 
        QFormer & LLaMA-13B & 117.7 & -- &  --  \\
        
        DreamLLM~\cite{dong2023dreamllm} & 
        Linear & Vicuna-7B & 115.4 & -- &  --   \\
        
        \rowcolor{gray!15}
        InternVL-G (ours) & 
        Cross-Attn & QLLaMA & \textbf{128.2} & 79.2 & 113.7 \\  
        
    \end{tabular}
\caption{\textbf{Comparison of zero-shot image captioning.}
QLLaMA inherently possesses promising zero-shot captioning capabilities thanks to its scaled-up parameters and datasets.
}
\label{tab: zs_cap}
\end{table}

\subsection{Vision-Language Benchmarks}
\label{sec:vision-language_benchmarks}

In this section, we evaluate the inherent capabilities of \modelname on various vision-language tasks.

\noindent\textbf{Zero-Shot Image Classification.}
We conduct thorough validation of the zero-shot image classification capability of \modelname-C.
As depicted in Table~\ref{subtable:in1k_variant}, \modelname-C attains leading performance on various ImageNet variants~\cite{deng2009imagenet,hendrycks2021imagenet_a,hendrycks2021imagenet_r,recht2019imagenetv2,wang2019imagenet_sketch} and ObjectNet~\cite{barbu2019objectnet}. 
Compared to EVA-02-CLIP-E+~\cite{sun2023evaclip}, it exhibits stronger robustness to distribution shift, manifesting in a more consistent accuracy across ImageNet variants.
Additionally, as shown in Table~\ref{subtable:multilingual_in1k}, our model showcases robust multilingual capabilities, outperforming competing models~\cite{chen2022altclip, openclip, yang2022cnclip, carlsson2022mclip} on the multilingual ImageNet-1K benchmark.

\noindent\textbf{Zero-Shot Video Classification.}
Following previous methods \cite{radford2021clip,sun2023evaclip,wang2023internvid}, we report the top-1 accuracy and the mean of top-1 and top-5 accuracy on Kinetics-400/600/700 \cite{carreira2017k400, carreira2018k600, carreira2019k700}. 
As shown in Table \ref{tab: zs_video}, when sampling only a single center frame in each video, our method achieves an average accuracy of 76.1\%, 75.5\%, and 67.5\% on the three datasets, surpassing EVA-02-CLIP-E+~\cite{sun2023evaclip} by +6.3, +6.2, and +4.1 points, respectively.
Additionally, when uniformly sampling 8 frames in each video, we obtain at least 3.3 points of improvement compared to the single-frame setting, outperforming ViCLIP~\cite{wang2023internvid} trained using web-scale video data.
In summary, \modelname-C exhibits remarkable generalization capabilities in video classification.

\noindent\textbf{Zero-Shot Image-Text Retrieval.}
\modelname exhibits a powerful multilingual image-text retrieval capability.
In Table~\ref{tab: clip zs retrieval}, we evaluate these capabilities in English using the Flickr30K~\cite{plummer2015flickr30k} and COCO~\cite{chen2015cococaption} datasets, as well as in Chinese using the Flickr30K-CN~\cite{lan2017flickrcn} and COCO-CN~\cite{li2019cococn}.
Additionally, we leverage the XTD dataset \cite{aggarwal2020xtd} to evaluate the multilingual image-text retrieval capability across 8 languages (see supplementary materials).
In summary, \modelname-C achieves state-of-the-art performance across most retrieval metrics, and with the second stage of pre-training, \modelname-G further enhances zero-shot image-text retrieval performance. 
These improvements in retrieval tasks suggest a more effective alignment between visual and linguistic features, through additional image encoding using the language middleware--QLLaMA.

\noindent\textbf{Zero-Shot Image Captioning.}
Benefiting from vision-language generative training on a vast collection of high-quality image-text pairs, our QLLaMA possesses promising capability in zero-shot image captioning.
As shown in Table~\ref{tab: zs_cap}, QLLaMA surpasses other models in zero-shot performance on the COCO Karpathy test set~\cite{chen2015cococaption}.
It also achieves comparable results to current state-of-the-art models on both the Flickr30K Karpathy test~\cite{plummer2015flickr30k} and the NoCaps val set~\cite{agrawal2019nocaps}. 
When InternVL is linked with an LLM (\eg, Vicuna-7B/13B \cite{zheng2023vicuna}) and subjected to SFT, a notable enhancement in zero-shot performance is observed for both Flickr30K and NoCaps, as shown in Table~\ref{tab:sota_results}.

\subsection{Multi-Modal Dialogue Benchmarks}

Beyond the traditional multi-modal tasks, the emergence of ChatGPT~\cite{openai2023gpt4} has led to a growing focus on evaluating the performance of multi-modal models in real usage scenarios, specifically within the realm of multi-modal dialogue. 
We conducted testing of \modelname-Chat models on two prominent multi-modal dialogue benchmarks, including MME~\cite{fu2023mme} and POPE~\cite{li2023pope}. 
MME is a comprehensive benchmark that includes 14 sub-tasks focusing on the model's perception and cognition capabilities. POPE is a popular dataset used to evaluate object hallucination.
As shown in Table~\ref{tab:sota_results}, it clearly demonstrates that our models exhibit superior performance compared with previous methods, under the condition of fair trainable parameter counts.

\subsection{Ablation Study}

\textbf{Hyperparameters of InternViT-6B.}
As discussed in Section \ref{sec:model_design}, we explored variations in model depth \{32, 48, 64, 80\}, head dimension \{64, 128\}, and MLP ratio \{4, 8\}, resulting in 16 distinct models.
In selecting the optimal model, we initially narrowed down our focus to 6 models, chosen based on their throughput, as listed in Table~\ref{tab:ablation_model_config}. 
These models underwent further evaluation using contrastive learning on a 100M subset of LAION-en~\cite{schuhmann2022laion5b} over 10K iterations.
For the experimental setup, the primary difference was the use of a randomly initialized text encoder from CLIP-L~\cite{radford2021clip}, in order to speed up the training.
For the sake of accuracy, inference speed, and training stability, we ultimately chose variant 3 as the final InternViT-6B.

\begin{table}[t]\scriptsize
\renewcommand{\arraystretch}{1.0}
\centering

\setlength\tabcolsep{2.9pt}
\begin{tabular}{l|cccc|ccc|c}
    name & width & depth & MLP & \#heads & \#param & FLOPs &  throughput & zs IN \\
    
    \hline
    
    variant 1 & 3968 & 32 & 15872 & 62 & 6051M & 1571G & 35.5 / 66.0 & 65.8 \\
    variant 2 & 3200 & 48 & 12800 & 50 & 5903M & 1536G & 28.1 / 64.9 & 66.1 \\
    \rowcolor{gray!15}
    variant 3 & 3200 & 48 & 12800 & 25 & 5903M & 1536G & 28.0 / 64.6 & 66.2 \\
    variant 4 & 2496 & 48 & 19968 & 39 & 5985M & 1553G & 28.3 / 65.3 & 65.9 \\
    variant 5 & 2816 & 64 & 11264 & 44 & 6095M & 1589G & 21.6 / 61.4 & 66.2 \\
    variant 6 & 2496 & 80 & 9984  & 39 & 5985M & 1564G & 16.9 / 60.1 & 66.2 \\
    
\end{tabular}
\caption{
\textbf{Comparison of hyperparameters in InternViT-6B.} 
The throughput (img/s) and GFLOPs are measured at 224$\times$224 input resolution, with a batch size of 1 or 128 on a single A100 GPU.
Flash Attention \cite{dao2022flashattention} and bf16 precision are used during testing.
``zs IN" denotes the zero-shot top-1 accuracy on the ImageNet-1K
validation set~\cite{deng2009imagenet}.
The final selected model is marked in \colorbox{baselinecolor}{gray}.
}
\label{tab:ablation_model_config}
\end{table}

\begin{table}[t]\scriptsize
\renewcommand{\arraystretch}{1.0}
\centering

\setlength\tabcolsep{1.05pt}
\begin{tabular}{lll|c|c|c|ccc}
    visual & glue & \multirow{2}{*}{LLM}  &  \multirow{2}{*}{dataset}   & dialogue & caption & \multicolumn{3}{c}{visual question answering}  \\
    encoder & layer &  &  & MME & NoCaps & OKVQA  & VizWiz$\rm_{val}$ & GQA  \\
    
    \hline
    
    EVA-E  & MLP & V-7B & 665K~\cite{liu2023improved} & 970.5 & 75.1 & 40.1 & 25.5 & 41.3 \\
    
    IViT-6B & MLP & V-7B & 665K~\cite{liu2023improved} & 1022.3 & 80.8 & 42.9 & 28.3  & 45.8 \\
    
    IViT-6B & QLLaMA & V-7B & 665K~\cite{liu2023improved} & 1227.5 & 94.5 & 51.0 & 38.4 & 57.4 \\

    \hline
    
    IViT-6B & QLLaMA & V-7B & Ours & 1298.5 & 120.5 & 51.8 & 44.9 & 57.7 \\

    IViT-6B & QLLaMA & V-13B & Ours & 1317.2 & 123.1 & 55.5 & 55.7 & 59.5 \\
    
    
\end{tabular}
\caption{
\textbf{Ablation studies of using InternVL to build multi-modal dialogue system.} 
V-7B and V-13B denote Vicuna-7B/13B \cite{zheng2023vicuna}, respectively.
``IViT-6B'' represents our InternViT-6B. 
}
\label{tab:ablation_component}
\end{table}

\noindent
\textbf{Consistency of Feature Representation.}
In this study, we validate the consistency of the feature representation of InternVL with off-the-shelf LLMs. 
We adopt a minimalist setting, \ie conducting a single-stage SFT using only the LLaVA-Mix-665K~\cite{li2021improved} dataset. Moreover, only the MLP layers are trainable, thereby confirming the inherent alignment level among features from various vision foundation models and LLMs. 
The results are shown in Table~\ref{tab:ablation_component}.
We observed that compared to EVA-E \cite{sun2023evaclip}, our InternViT-6B achieves better performance under this simple setup. 
Additionally, it is noteworthy that performance across all three tasks saw significant improvement when using QLLaMA as the ``glue layer".
These significant improvements clearly delineate that \emph{the feature representation of InternVL is more consistent with the off-the-shelf LLM.}

\section{Conclusion}
\label{sec:conclusion}

In this paper, we present \modelname, a large-scale vision-language foundation model that scales up the vision foundation model to 6 billion parameters and is aligned for generic visual-linguistic tasks. 
Specifically, we design a large-scale vision foundation model InternViT-6B, progressively align it with an LLM-initialized language middleware QLLaMA, and leverage web-scale image-text data from various sources for efficient training. 
It bridges the gap between vision foundation models and LLMs, and demonstrates proficiency in a wide range of generic visual-linguistic tasks, such as image/video classification, image/video-text retrieval, image captioning, visual question answering, and multi-modal dialogue. 
We hope this work could contribute to the development of the VLLM community.

\section*{Acknowledgement}
We thank Shenglong Zhang, Beitong Zhou, Xinyue Zhang, Dongxing Shi, Weigao Sun, Xingcheng Zhang, and Zhifeng Yue for their contributions to the optimization of the training framework. 
We thank Zhenhang Huang for his assistance in data preparation.

\clearpage
\newpage

\appendix

\section{Supplementary Materials}

\subsection{More Experiments}

\noindent
\textbf{Zero-Shot Image Classification on 20 Datasets.}
In this section, we expand our examination to showcase the effectiveness and robustness of InternVL in 20 different zero-shot image classification benchmarks. 
As indicated in Table \ref{tab: clip_zs_img_cls_20}, InternVL registers an average performance of 78.1\% across all 20 benchmarks. 
This performance notably exceeds that of the previously leading method, EVA-02-CLIP-E+~\cite{fang2023eva02}, by a margin of 1.0 points. 
This underscores that, beyond ImageNet~\cite{deng2009imagenet} and its variants, InternVL possesses robust generalization capabilities across a variety of different domains in zero-shot image classification.

\noindent
\textbf{Zero-Shot Image-Text Retrieval on XTD.}
Table \ref{tab: zs_xtd} reports the results of InternVL on the multilingual image-text retrieval dataset XTD \cite{aggarwal2020xtd}, spanning eight languages.
As can be seen, InternVL-C achieves an average recall@10 score of 95.1\% across these languages. The second stage model, InternVL-G, further improves retrieval performance. It attains the highest scores in each individual language and establishes a new record for average performance at 96.6\%.

\noindent
\textbf{Zero-Shot Video Retrieval.}
In Table \ref{tab: video zs retrieval}, we present our results of zero-shot video-text retrieval on the MSR-VTT dataset~\cite{xu2016msrvtt} using our InternVL models, \ie InternVL-C and InternVL-G. 
In the 1-frame setting, we select a single central frame from each video. In the 8-frame setting, we uniformly extract 8 frames from each video, treat them as independent images for encoding, and then average the embeddings.
The results showcase consistent improvement across various metrics such as R@1, R@5, R@10, and the average score. 
Importantly, both models exhibit promising outcomes in single-frame and multi-frame configurations, with InternVL-G achieving slightly higher performance than InternVL-C, especially in the multi-frame setting. 
These results underscore the effectiveness of QLLaMA in harmonizing visual and linguistic features.

\noindent
\textbf{Fine-tuned Image-Text Retrieval.}
In Table \ref{tab: finetune retrieval}, we report the fine-tuned image-text retrieval results of InternVL, on both the English and Chinese versions of the Flickr30K dataset~\cite{plummer2015flickr30k, lan2017flickrcn}. 
The specific hyperparameters for fine-tuning are shown in Table \ref{tab:train_cfg_ft_retrieval}.
As can be seen, our models obtain competitive performance, with InternVL-G-FT marginally surpassing InternVL-C-FT in both datasets. 
Notably, in the highly challenging Flickr30K-CN, both models show a promising ability to handle cross-lingual retrieval tasks.
These results demonstrate the effectiveness of our language middleware, especially in the retrieval tasks.

\noindent
\textbf{Tiny LVLM.}
Tiny LVLM \cite{shao2023tiny} is an ability-level benchmark for evaluating the performance of multimodal dialogue models.
It provides a systematic assessment of five categories of multimodal capabilities, including visual perception, visual knowledge acquisition, visual reasoning, visual commonsense, and object hallucination.
We report our results on Tiny LVLM in Table \ref{tab: tiny_lvlm}. 

\subsection{More Ablation Studies}


\begin{table}[t]
\scriptsize
\renewcommand{\arraystretch}{1.0}
    \centering

    \setlength\tabcolsep{2.4pt}
    \begin{tabular}{l|cccccccc|c}
        method & EN & ES & FR & ZH & IT & KO & RU & JP & avg. \\
        \hline
        
        mUSE m3~\cite{yang2020muse} &
        85.3 & 78.9 & 78.9 & 76.7 & 73.6 & 67.8 & 76.1 & 70.7 & 76.0 \\
        
        M-CLIP~\cite{carlsson2022mclip} &
        92.4 & 91.0 & 90.0 & 89.7 & 91.1 & 85.2 & 85.8 & 81.9 & 88.4 \\
        
        MURAL~\cite{jain2021mural} & 
        $-$ & 92.9 & $-$ & 89.7 & 91.8 & 88.1 & 87.2 & $-$ & $-$\\

        AltCLIP~\cite{chen2022altclip} & 
        95.4 & 94.1 & 92.9 & 95.1 & 94.2 & 94.4 & 91.8 & 91.7 & 93.7 \\
        
        OpenCLIP-XLM-R-B~\cite{openclip} &
        95.8 & 94.4 & 92.5 & 91.8 & 94.4 & 86.3 & 89.9 & 90.7 & 92.0 \\

        OpenCLIP-XLM-R-H~\cite{openclip} &
        97.3 & 96.1 & 94.5 & 94.7 & 96.0 & 90.2 & 93.9 & 94.0 & 94.6 \\
        
        \rowcolor{gray!15}
        InternVL-C (ours) & 97.3 & 95.7 & 95.1 & 95.6 & 96.0 & 92.2 & 93.3 & 95.5 & 95.1 \\  
        \rowcolor{gray!15}
        InternVL-G (ours) & \textbf{98.6} & \textbf{97.7} & \textbf{96.5} & \textbf{96.7} & \textbf{96.9} & \textbf{95.1} & \textbf{94.8} & \textbf{96.1} & \textbf{96.6}  \\  
         
    \end{tabular}
    \caption{\textbf{Comparison of zero-shot multilingual image-text retrieval performance on the XTD dataset.}
    Multiple languages include English (EN), Spanish (ES), French (FR), Chinese (ZH), Italian (IT), Korean (KO), Russian (RU), and Japanese (JP).
    We follow M-CLIP~\cite{carlsson2022mclip} to report the recall@10 on Image-to-Text. 
}
\label{tab: zs_xtd}
\end{table}

\begin{table}[t]\scriptsize
\renewcommand{\arraystretch}{1.0}
\setlength\tabcolsep{3.1pt}
\begin{tabular}{l|c|ccc|ccc|c}
    & & \multicolumn{6}{c|}{MSR-VTT (1K test set)~\cite{xu2016msrvtt}} & \\
    & & \multicolumn{3}{c|}{Video $\rightarrow$ Text} & \multicolumn{3}{c|}{Text $\rightarrow$ Video}  \\
    \multirow{-3}{*}{method} & \multirow{-3}{*}{\#F} & R@1 &  R@5 &  R@10 &  R@1 &  R@5 &  R@10 & \multirow{-3}{*}{avg.}  \\
    \hline

     OpenAI CLIP-L~\cite{radford2021clip} & 1 &
     27.8 & 49.4 & 58.0 & 29.0 & 50.5 & 59.2 & 45.7 
     \\

    \rowcolor{gray!15}
     InternVL-C (ours)  & 1 & 35.3 & 56.6 & 66.6 & 37.5 & 60.9 & \textbf{70.9} & 54.6
     \\  
     
     \rowcolor{gray!15}
     InternVL-G (ours) & 1 & \textbf{36.6} & \textbf{58.3} & \textbf{67.7} & \textbf{39.1} & \textbf{61.7} & 70.7 & \textbf{55.7} 
     \\
     \hline

     OpenAI CLIP-L~\cite{radford2021clip} & 8 &
     26.6 & 50.8 & 61.8 & 30.7 & 54.4 & 64.0 & 48.1
     \\

     Florence~\cite{yuan2021florence}   & 8 & 
     --   & --   & --   & 37.6 & 63.8 & 72.6 & --
     \\

     InternVideo$^\dagger$~\cite{wang2022internvideo}  & 8 & 
     39.6 & -- & -- & 40.7 & -- & -- & -- 
     \\

     UMT-L$^\dagger$~\cite{li2023unmasked}  & 8 & 
     38.6 & 59.8 & 69.6 & 42.6 & 64.4 & 73.1 & 58.0
     \\ 
     
     LanguageBind$^\dagger$~\cite{zhu2023languagebind}  & 8 & 
     40.9 & 66.4 & 75.7 & 44.8 & 70.0 & 78.7 & 62.8
     \\

     \rowcolor{gray!15}
     InternVL-C (ours)  & 8 & 40.2 & 63.1 & 74.1 & 44.7 & 68.2 & 78.4 & 61.5
     \\  
     
     \rowcolor{gray!15}
     InternVL-G (ours) & 8 & \textbf{42.4} & \textbf{65.9} & \textbf{75.4} & \textbf{46.3} & \textbf{70.5} & \textbf{79.6} & \textbf{63.4} 
     \\
     
\end{tabular}
\caption{\textbf{Comparison of zero-shot video-text retrieval performance on MSR-VTT.}
``\#F" denotes the number of frames.
$^\dagger$~These models are trained with temporal attention layers.
}
\label{tab: video zs retrieval}
\end{table}

\begin{table}[t]\scriptsize
\renewcommand{\arraystretch}{1.0}
\begin{subtable}{1\textwidth}
    \setlength\tabcolsep{3.9pt}
    \begin{tabular}{l|ccc|ccc|c}
        &  \multicolumn{6}{c|}{Flickr30K (English, 1K test set)~\cite{plummer2015flickr30k}} & \\
         & \multicolumn{3}{c|}{Image $\rightarrow$ Text} & \multicolumn{3}{c|}{Text $\rightarrow$ Image}  \\
        \multirow{-3}{*}{method} & R@1 &  R@5 &  R@10 &  R@1 &  R@5 &  R@10 & \multirow{-3}{*}{avg.}  \\
        \hline

         ALIGN~\cite{jia2021scaling} &
         95.3 & 99.8 & 100.0 & 84.9 & 97.4 & 98.6 & 96.0 \\
         
         FILIP~\cite{yao2021filip}&
         96.6 & 100.0 & 100.0 & 87.1 & 97.7 & 99.1 & 96.8 \\
         
         Florence~\cite{yuan2021florence} &
         97.2 & 99.9 & $-$ & 87.9 & 98.1 & $-$ & $-$ \\

         BLIP~\cite{li2022blip} & 97.4 & 99.8 & 99.9 & 87.6 & 97.7 & 99.0 & 96.9 \\
         
         OmniVL~\cite{wang2022omnivl}  &
         97.3 & 99.9 & 100.0 & 87.9 & 97.8 & 99.1 & 97.0 \\

         BEiT-3~\cite{wang2023beit3} & 
         97.5 & 99.9 & 100.0 & 89.1 & 98.6 & \textbf{99.3} & 97.4 \\

         ONE-PEACE~\citep{wang2023onepeace} & 97.6 & 100.0 & 100.0 & 89.6 & 98.0 & 99.1 & 97.4 \\   

         \rowcolor{gray!15}
         InternVL-C-FT (ours)   & 
         97.2 & 100.0 & 100.0 & 88.5 & 98.4 & 99.2 & 97.2 \\  
         
         \rowcolor{gray!15}
         InternVL-G-FT (ours)  & \textbf{97.9} & \textbf{100.0} & \textbf{100.0} & \textbf{89.6} & \textbf{98.6} & 99.2 & \textbf{97.6} \\
         \multicolumn{7}{c}{}\\
         method & \multicolumn{6}{c|}{Flickr30K-CN (Chinese, 1K test set)~\citep{lan2017flickrcn}} & avg. \\
         
        \hline
         Wukong-ViT-L~\citep{gu2022wukong} & 
         92.7 & 99.1 & 99.6 & 77.4 & 94.5 & 97.0 & 93.4 \\

         CN-CLIP-ViT-H~\citep{yang2022cnclip}  & 
         95.3 & 99.7 & 100.0 & 83.8 & 96.9 & 98.6 & 95.7  \\ 

         R2D2-ViT-L~\citep{xie2022zero}  & 
         95.6 & 99.8 & 100.0 & 84.4 & 96.7 & 98.4 & 95.8  \\
        
         \rowcolor{gray!15}
         InternVL-C-FT (ours)  & 96.5 & 99.9 & 100.0 & 85.2 & 97.0 & 98.5 & 96.2
         \\  
         
         \rowcolor{gray!15}
         InternVL-G-FT (ours) & \textbf{96.9} & \textbf{99.9} & \textbf{100.0} & \textbf{85.9} & \textbf{97.1} & \textbf{98.7} & \textbf{96.4}
         \\  
    \end{tabular}
\end{subtable}
\caption{\textbf{Comparison of fine-tuned image-text retrieval performance.}
We evaluate English and Chinese image-text retrieval using Flickr30K~\cite{plummer2015flickr30k} and Flickr30K-CN~\cite{lan2017flickrcn}, with separate fine-tuning for each to prevent data leakage.
}
\label{tab: finetune retrieval}
\end{table}

\begin{table*}[ht]\scriptsize
\centering
\setlength\tabcolsep{3.85pt}
\renewcommand{\arraystretch}{1.0}
    \begin{tabular}{l|cccccccccccccccccccc|c}
        method
        &
        \rotatebox[origin=l]{90}{{CIFAR-10~\cite{krizhevsky2009cifar}}} &
        \rotatebox[origin=l]{90}{{CIFAR-100~\cite{krizhevsky2009cifar}}} & 
        \rotatebox[origin=l]{90}{{MNIST~\cite{lecun1998mnist}}} & 
        \rotatebox[origin=l]{90}{{Caltech-101~\cite{fei2004learning}}} & 
        \rotatebox[origin=l]{90}{{SUN397~\cite{xiao2010sun}}} & 
        \rotatebox[origin=l]{90}{{FGVC Aircraft~\cite{maji2013fgvc}}} & 
        \rotatebox[origin=l]{90}{{Country-211~\cite{radford2021clip}}} & 
        \rotatebox[origin=l]{90}{{Stanford Cars~\cite{krause2013cars}}} &
        \rotatebox[origin=l]{90}{{Birdsnap~\cite{berg2014birdsnap}}} & 
        \rotatebox[origin=l]{90}{{DTD~\cite{cimpoi2014d2d}}} & 
        \rotatebox[origin=l]{90}{{Eurosat~\cite{helber2019eurosat}}} & 
        \rotatebox[origin=l]{90}{{FER2013~\cite{goodfellow2013fer2013}}} & 
        \rotatebox[origin=l]{90}{{Flowers-102~\cite{nilsback2008flowers}}} & 
        \rotatebox[origin=l]{90}{{Food-101~\cite{bossard2014food101}}} & 
        \rotatebox[origin=l]{90}{{GTSRB~\cite{stallkamp2012gtsrb}}} & 
        \rotatebox[origin=l]{90}{{Pets~\cite{parkhi2012cats}}} & 
        \rotatebox[origin=l]{90}{{Rendered SST2~\cite{radford2021clip}}} & 
        \rotatebox[origin=l]{90}{{Resisc45~\cite{cheng2017resisc45}}} & 
        \rotatebox[origin=l]{90}{{STL10~\cite{coates2011stl10}}} & 
        \rotatebox[origin=l]{90}{{VOC2007~\cite{everingham2015pascal}}} &
        \rotatebox[origin=l]{90}{avg. top-1 acc.}
        \\
        \hline
         OpenAI CLIP-L+~\cite{radford2021clip} & 
         94.9 & 74.4 & 79.0 & 87.2 & 68.7 & 33.4 & 34.5 & 79.3 & 41.0 & 56.0 & 61.5 & 
         49.1 & 78.6 & 93.9 & 52.4 & 93.8 & 70.7 & 65.4 & 99.4 & 78.1 & 69.6 \\
         
         EVA-01-CLIP-g~\cite{sun2023evaclip} &
         98.3 & 88.7 & 62.3 & 87.7 & 74.2 & 32.4 & 28.6 & 91.7 & 50.0 & 61.3 & 73.6 & 
         52.2 & 74.5 & 93.5 & 49.1 & 94.2 & 58.4 & 70.3 & 98.9 & 83.2 & 71.2 \\
         
         OpenCLIP-g~\citep{openclip} & 
         98.2 & 84.7 & 71.9 & 88.1 & 74.1 & 44.6 & 30.9 & 94.0 & 51.0 & 68.7 & 64.7 & 
         55.8 & 81.0 & 92.4 & 49.7 & 93.9 & 56.7 & 69.6 & 98.9 & 81.6 & 72.5 \\
        
         OpenCLIP-H~\citep{openclip} &
         97.4 & 84.7 & 72.9 & 85.0 & 75.2 & 42.8 & 30.0 & 93.5 & 52.9 & 67.8 & 72.7 &
         52.0 & 80.1 & 92.7 & 58.4 & 94.5 & 64.3 & 70.5 & 98.5 & 77.7 & 73.2 \\

         EVA-02-CLIP-L+~\cite{sun2023evaclip} & 
         98.9 & 89.8 & 64.3 & 89.5 & 74.8 & 37.5 & 33.6 & 91.6 & 45.8 & 64.5 & 71.4 &
         51.0 & 77.2 & 94.2 & 57.6 & 94.2 & 64.6 & 69.8 & \textbf{99.7} & 82.7 & 72.6 \\

         EVA-01-CLIP-g+~\cite{sun2023evaclip} &
         99.1 & 90.1 & 71.8 & 88.1 & 74.3 & 39.4 & 30.8 & 90.7 & 52.6 & 67.3 & 73.2 &
         56.0 & 79.7 & 93.7 & 66.5 & 94.8 & 58.6 & 71.4 & 99.5 & 82.9 & 74.0 \\
         
         OpenCLIP-G~\citep{openclip} & 
         98.2 & 87.5 & 71.6 & 86.4 & 74.5 & 49.7 & 33.8 & 94.5 & 54.5 & 69.0 & 70.0
         & \textbf{59.5} & 81.5 & 93.1 & 62.5 & 95.2 & 65.2 & 72.6 & 98.5 & 80.7 & 74.9 \\

         EVA-02-CLIP-E~\cite{sun2023evaclip} & 
         99.3 & 92.5 & 76.7 & 89.0 & \textbf{76.5} & 47.9 & 34.7 & 94.4 & 56.3 & 68.2 & 77.6 
         & 55.1 & 82.5 & 95.2 & 67.1 & 95.6 & 61.1 & 73.5 & 99.2 & 83.0 & 76.3 \\

         EVA-02-CLIP-E+~\cite{sun2023evaclip} & 
         99.3 & 93.1 & 74.7 & \textbf{90.5} & 75.1 & \textbf{54.1} & \textbf{35.7} & \textbf{94.6} & 58.1 & 68.2 & 75.8
         & 58.6 & 84.5 & 94.9 & \textbf{67.7} & 95.8 & 61.4 & \textbf{75.6} & 99.2 & \textbf{85.6} & 77.1 \\

         \rowcolor{gray!15}
         InternVL-C (ours) & \textbf{99.4} & \textbf{93.2} & \textbf{80.6} & 89.5 & 76.0 & 52.7 & 34.1 & 94.2 & \textbf{72.0} & \textbf{70.7} & \textbf{79.4} & 56.2 & \textbf{86.1} & \textbf{95.3} & 65.5 & \textbf{96.0} & \textbf{67.9} & 74.2 & 99.5 & 80.0 & \textbf{78.1} \\
        \end{tabular}
\caption{\textbf{Comparison of zero-shot image classification performance on 20 other datasets.}
These results indicate that, in addition to ImageNet \cite{deng2009imagenet}, InternVL also possesses good generalization capabilities in zero-shot image classification across various domains.
}
\label{tab: clip_zs_img_cls_20}
\end{table*}

\begin{table}[t]
\scriptsize
\renewcommand{\arraystretch}{1.0}
    \centering

    \setlength\tabcolsep{1.95pt}
    \begin{tabular}{ll|ccccc|c}
        method & LLM & VR & VP & VKA & VC & OH & Overall \\
        \hline
        MiniGPT-4~\cite{zhu2023minigpt4} & Vicuna-7B & 37.6 & 37.8 & 17.6 & 49.0 & 50.7 & 192.6 \\
        LLaVA~\cite{liu2023llava} & Vicuna-7B & 41.6 & 38.3 & 18.7 & 49.4 & 49.0 & 197.0 \\
        VisualGLM~\cite{du2022glm} & ChatGLM-6B & 37.3 & 36.3 & 46.9 & 37.6 & 54.0 & 211.9 \\
        Otter~\cite{li2023otter} & Otter-9B & 41.6 & 37.0 & 15.1 & 52.4 & 74.0 & 216.4 \\
        LLaMA-Adapter-V2~\cite{gao2023llama-adapterv2} & LLaMA-7B & 43.5 & 46.8 & 22.3 & 56.0 & 60.7 & 229.2 \\
        Lynx~\cite{zeng2023lynx} & Vicuna-7B & 52.2 & 65.8 & 17.6 & 57.4 & 86.3 & 279.2 \\
        BLIP-2~\cite{li2023blip2} & FlanT5xl & 44.9 & 49.0 & 64.1 & 44.0 & 82.7 & 284.7 \\
        InstructBLIP~\cite{instructblip} & Vicuna-7B & 46.7 & 48.0 & 61.7 & 59.2 & 85.0 & 300.6 \\
        LLaVA-1.5~\cite{liu2023improved} & Vicuna-7B & 55.6 & 49.0 & 57.0 & 57.2 & 88.3 & 307.2 \\
        Qwen-VL-Chat~\cite{bai2023qwenvl} & Qwen-7B & 62.4 & 54.5 & 55.1 & 54.8 & 90.0 & 316.8 \\
        Bard~\cite{google_bard} & Bard & 64.2 & 57.0 & 68.1 & 59.6 & 70.7 & 319.6 \\
        InternLM-XComposer~\cite{zhang2023internlmxcomposer} & InternLM-7B & 55.8 & 53.8 & 64.1 & 61.8 & 87.0 & 322.5 \\
        \rowcolor{gray!15}
        InternVL-Chat (ours) & Vicuna-13B & 56.4 & 52.3 & 68.0 & 62.0 & 89.0 & \textbf{327.6}  \\  
         
    \end{tabular}
    \caption{\textbf{Evaluation of Tiny LVLM test set.}
    Here we report five categories of multimodal capabilities, including visual reasoning (VR), visual perception (VP), visual knowledge acquisition (VKA), visual commonsense (VC), and object hallucination (OH).
}
\label{tab: tiny_lvlm}
\end{table}

\noindent
\textbf{Compatibility with Other LLM.} 
In this experiment, we test the compatibility of InternVL with LLMs other than Vicuna \cite{zheng2023vicuna}. The experimental setup used here is the same as in Table \ref{tab:sota_results} of the main paper. As shown in Table \ref{tab:compatibility_with_other_llm}, InternLM-7B \cite{2023internlm} achieves slightly better performance than Vicuna-7B \cite{zheng2023vicuna}. This indicates that our InternVL exhibits promising compatibility with various LLMs.

\noindent
\textbf{Efficiency Analysis.}
In this study, we analyze the computational efficiency of InternVL in encoding image-text pairs. The entire encoding process consists of two parts: image encoding and text encoding. The analysis covered two models (InternVL-C and InternVL-G) and their performance across three different image sizes (224, 336, and 448).
The results are shown in Table \ref{tab:ablation_efficiency}.

From these results, we find that:
(1) As the image size increases, the encoding time also significantly increases, leading directly to a decrease in frame rate;
(2) InternVL-G slightly increased the encoding time due to the introduction of QLLaMA for secondary image encoding, but it still maintains a reasonable frame rate across all image sizes;
(3) Even though we scale up the text encoder, the additional cost of text encoding is not significant, as the main time expenditure lies in image encoding.
In summary, when choosing between InternVL-C and InternVL-G, one should weigh the trade-off between computational efficiency and potential performance improvements based on specific requirements.
Additionally, these results were measured using PyTorch with Flash Attention \cite{dao2022flashattention} and bf16 precision, and there is still considerable room for optimization, such as using model quantization and TensorRT.

\begin{table}[t]
\scriptsize
\renewcommand{\arraystretch}{1.0}
    \centering
    \setlength\tabcolsep{2.5pt}
    \begin{tabular}{lll|cccc|cc}
        visual  & glue  &                       & \multicolumn{4}{c|}{visual question answering} &  \multicolumn{2}{c}{dialogue}  \\
        encoder & layer & \multirow{-2}{*}{LLM} & VQA$^\text{v2}$ & GQA & VizWiz & VQA$^\text{T}$ & MME & POPE \\
        
        \hline

        IViT-6B & MLP & Vicuna-7B   & 79.3 & 62.9 & 52.5 & 57.0 & 1525.1 & 86.4 \\
        IViT-6B & MLP & InternLM-7B & 79.7 & 63.2 & 53.1 & 58.0 & 1532.8 & 86.4 \\
        
    \end{tabular}
    \caption{\textbf{Compatibility with other LLM.}
    Here we use InternLM \cite{2023internlm} as an example to verify the compatibility of InternVL with LLMs other than Vicuna \cite{zheng2023vicuna}.
    The experimental settings used here are the same as in Table \ref{tab:sota_results} of the main paper.
}
\label{tab:compatibility_with_other_llm}
\end{table}

\begin{table}[t]
\scriptsize
\renewcommand{\arraystretch}{1.0}
    \centering

    \setlength\tabcolsep{3.1pt}
    \begin{tabular}{l|c|cc|c|c|c}
               & image  & \multicolumn{2}{c|}{encode image (ms)}& \multicolumn{1}{c|}{encode text (ms)} & total & \\
        \multirow{-2}{*}{method}  & size  & InternViT-6B & QLLaMA & QLLaMA & time &\multirow{-2}{*}{FPS} \\
        
        \hline

        InternVL-C & 224 & 15.5 & -- & 4.9 & 20.4 & 48.9 \\
        InternVL-C & 336 & 35.2 & -- & 4.9 & 40.1 & 24.9 \\
        InternVL-C & 448 & 66.9 & -- & 4.9 & 71.8 & 13.9 \\
        \hline
        InternVL-G & 224 & 15.5 & 8.2 & 4.9 & 28.6 & 35.0 \\
        InternVL-G & 336 & 35.2 & 10.3 & 4.9 & 50.4 & 19.8 \\
        InternVL-G & 448 & 66.9 & 12.8 & 4.9 & 84.6 & 11.8 \\
        
    \end{tabular}
    \caption{\textbf{Efficiency analysis of InternVL for encoding image-text pairs.}
    The total time to encode an image-text pair includes both the image encoding part and the text encoding part.
    We measure the time cost with a batch size of 128 on a single A100 GPU.
    Flash Attention \cite{dao2022flashattention} and bf16 precision are used during testing.
}
\label{tab:ablation_efficiency}
\end{table}

\subsection{Detailed Training Settings}
\noindent
\textbf{Settings of Stage~1.}
As shown in Table~\ref{tab:train_cfg_stage1_stage2}, in this stage, the image encoder InternViT-6B is randomly initialized using the BEiT's initialization method \cite{bao2021beit}, and the text encoder LLaMA-7B is initialized with the pre-trained weights from \cite{cui2023chinesellama}, a multilingual LLaMA-7B.
All parameters are fully trainable. 
We employ the AdamW optimizer~\cite{loshchilov2017adamw} with $\beta_{1}=0.9$, $\beta_{2}=0.95$, weight decay at 0.1, and a cosine learning rate schedule starting at 1e-3 and 1e-4 for the image and text encoders, respectively.
We adopt a uniform drop path rate of 0.2.
The training involves a total batch size of 164K across 640 A100 GPUs, extending over 175K iterations to process about 28.7 billion samples.
To enhance efficiency, we initially train at a 196$\times$196 resolution, masking 50\% of image tokens~\cite{li2023flip}, and later switch to 224$\times$224 resolution without masking for the final 0.5 billion samples.

\noindent
\textbf{Settings of Stage 2.}
In this stage, InternViT-6B and QLLaMA inherit their weights from the first stage, while the learnable queries and cross-attention layers in QLLaMA are randomly initialized.
Benefiting from the powerful encoding capabilities learned in the first stage, we keep both InternViT-6B and QLLaMA frozen and only train the newly added parameters. 
The input images are processed at a resolution of 224$\times$224.
For optimization, the AdamW optimizer~\cite{loshchilov2017adamw} is employed with $\beta_{1}=0.9$, $\beta_{2}=0.98$, weight decay set at 0.05, and a total batch size of 20K.
The training extends over 80K steps across 160 A100 GPUs, inclusive of 2K warm-up steps, and is governed by a cosine learning rate schedule with a peak learning rate of 5e-5.
More detailed training settings are listed in Table~\ref{tab:train_cfg_stage1_stage2}.

\begin{table}[t]
\scriptsize
\renewcommand{\arraystretch}{1.0}
    \centering

    \setlength\tabcolsep{4pt}
    \begin{tabular}{l|cc}
        config & stage 1 & stage 2 \\
        \hline

        image enc. weight init. & random init.~\cite{bao2021beit} & from stage 1  \\
        text enc. weight init. & from~\cite{cui2023chinesellama} & from stage 1  \\
        image enc. peak learning rate & 1e-3 & frozen \\
        text enc. peak learning rate & 1e-4 & frozen \\
        cross attn peak learning rate & -- & 5e-5 \\
        learning rate schedule & cosine decay & cosine decay \\
        optimizer & AdamW~\cite{loshchilov2017adamw} & AdamW~\cite{loshchilov2017adamw} \\
        optimizer hyper-parameters & $\beta_{1}$, $\beta_{2}$ = 0.9, 0.95 & $\beta_{1}$, $\beta_{2}$ = 0.9, 0.98 \\
         weight decay & 0.1 & 0.05 \\
         input resolution & 196$^2$ $\rightarrow$ 224$^2$ & 224$^2$ \\
         patch size & 14 & 14 \\
         total batch size & 164K & 20K \\
         warm-up iterations & 5K & 2K \\
         total iterations & 175K & 80K \\
         samples seen & 28.7B & 1.6B \\
         drop path rate~\cite{huang2016droppath} & uniform (0.2) & 0.0 \\
         data augmentation & random resized crop & random resized crop \\
         numerical precision & DeepSpeed bf16~\cite{rasley2020deepspeed} & DeepSpeed bf16~\cite{rasley2020deepspeed} \\
          trainable / total parameters & 13B / 13B & 1B / 14B \\
          GPUs for training & 640$\times$A100 (80G) & 160$\times$A100 (80G) \\
    \end{tabular}
    \caption{\textbf{Training settings of InternVL's stage 1 and stage 2.} 
    ``196$^2$ $\rightarrow$ 224$^2$" means we initially train at a 196$\times$196 resolution, and later switch to 224$\times$224 resolution for the final 0.5 billion samples, for higher training efficiency.
}
\label{tab:train_cfg_stage1_stage2}
\end{table}

\begin{table}[t]
\scriptsize
\renewcommand{\arraystretch}{1.0}
    \centering
    \setlength\tabcolsep{4pt}
    \begin{tabular}{l|c}
        config & retrieval fine-tuning  \\
        \hline
        image-text data & ~~~~~~~Flickr30K \cite{plummer2015flickr30k} / Flickr30K-CN \cite{lan2017flickrcn}~~~~~~~ \\
        peak learning rate & 1e-6 \\
        layer-wise lr decay rate & InternViT-6B (0.9), QLLaMA (0.9) \\
        learning rate schedule & cosine decay \\
        optimizer & AdamW~\cite{loshchilov2017adamw}  \\
        optimizer hyper-parameters & $\beta_{1}$, $\beta_{2}$ = 0.9, 0.999\\
         weight decay & 0.05 \\
         input resolution & 364$^2$ \\
         patch size & 14  \\
         total batch size & 1024 \\
         warm-up iterations & 100 \\
         training epochs & 10 \\
         drop path rate~\cite{huang2016droppath} & 0.3  \\
         data augmentation & random resized crop \& flip \\
         numerical precision & DeepSpeed bf16~\cite{rasley2020deepspeed}\\
          trainable / total parameters~~~~~~ & 14B / 14B \\
          GPUs for training & 32$\times$A100 (80G) \\
          
    \end{tabular}
    \caption{\textbf{Training settings of retrieval fine-tuning.} 
    We fine-tune InternVL on Flickr30K and Flickr30K-CN separately.
}
\label{tab:train_cfg_ft_retrieval}
\end{table}

\noindent
\textbf{Settings of Stage 3.}
At this stage, we have two different configurations.
One is to use InternViT-6B separately, as shown in Figure \ref{fig:inference_mode} (c).
The other is to use the entire InternVL model simultaneously, as shown in Figure~\ref{fig:inference_mode} (d). 

(1) InternVL-Chat (w/o QLLaMA): For this setup, we follow the training recipes of LLaVA-1.5 \cite{liu2023improved}. We use the same hyperparameters and datasets for supervised fine-tuning, \ie we first train the MLP layers with the LGS-558K \cite{liu2023llava} dataset, and then train the LLM with the LLaVA-Mix-665K \cite{liu2023improved} dataset, both for one epoch.
    
{(2) InternVL-Chat (w/ QLLaMA)}: For this more advanced setup, we also conducted the training in two steps. We first train the MLP layers with our custom SFT dataset and then fine-tune the LLM with it. Due to the expansion of the dataset, we increased the batch size to 512.

\noindent
\textbf{Settings of Retrieval Fine-tuning.}
In this experiment, all parameters of InternVL are set to be trainable. We conduct separate fine-tuning on the Flickr30K \cite{plummer2015flickr30k} and Flickr30K-CN \cite{lan2017flickrcn}. Following common practice \cite{li2023blip2}, a 364$\times$364 resolution is adopted for fine-tuning.
To avoid over-fitting, we apply a layer-wise learning rate decay of 0.9 to both InternViT-6B and QLLaMA, along with a drop path rate of 0.3 for InternViT-6B. 
The AdamW optimizer \cite{loshchilov2017adamw} is utilized, with a total batch size of 1024, for fine-tuning the InternVL model across 10 epochs. For more detailed training settings, please refer to Table \ref{tab:train_cfg_ft_retrieval}.

\begin{table}[t]
\scriptsize
\renewcommand{\arraystretch}{1.0}
    \centering

    \setlength\tabcolsep{4pt}
    \begin{tabular}{l|c}
        config & ImageNet linear probing  \\
        \hline
        peak learning rate & 0.2 \\
        learning rate schedule~~~~~~ & cosine decay \\
        optimizer & SGD  \\
        optimizer momentum & 0.9\\
         weight decay & 0.0 \\
         input resolution & 224$^2$ \\
         patch size & 14  \\
         total batch size & 1024 \\
         warm-up epochs & 1 \\
         training epochs & 10 \\
         data augmentation & ~~~~~~~~~random resized crop \& flip~~~~~~~~~ \\
         GPUs for training & 8$\times$A100 (80G) \\
    \end{tabular}
    \caption{\textbf{Training settings of ImageNet linear probing.} 
}
\label{tab:train_cfg_in_linear}
\end{table}
\begin{table}[t]
\scriptsize
\renewcommand{\arraystretch}{1.0}
    \centering

    \setlength\tabcolsep{6pt}
    \begin{tabular}{l|c}
        config & linear probing / head tuning / full tuning  \\
        \hline
        peak learning rate & 4e-5  \\
        layer-wise lr decay rate & -- / -- / 0.95 \\
        learning rate schedule & polynomial decay \\
        optimizer & AdamW \cite{loshchilov2017adamw} \\
        optimizer hyper-parameters & $\beta_{1}$, $\beta_{2}$ = 0.9, 0.999\\
         weight decay & 0.0 / 0.05 / 0.05 \\
         input resolution & 504$^2$ \\
         patch size & 14  \\
         total batch size & 16 \\
         warm-up iterations & 1.5K \\
         total iterations & 80K \\
         drop path rate~\cite{huang2016droppath} & 0.0 / 0.0 / 0.4 \\
         data augmentation & default augmentation in MMSeg \cite{contributors2020mmsegmentation} \\
          numerical precision & DeepSpeed bf16~\cite{rasley2020deepspeed}\\
         GPUs for training & 8$\times$A100 (80G) \\
    \end{tabular}
    \caption{\textbf{Training settings of ADE20K semantic segmentation.} 
    We list the hyperparameters for three different configurations, including linear probing, head tuning, and full-parameter tuning.
}
\label{tab:train_cfg_ade20k}
\end{table}

\begin{figure*}[t!]
    \centering
    \includegraphics[width=1\textwidth]{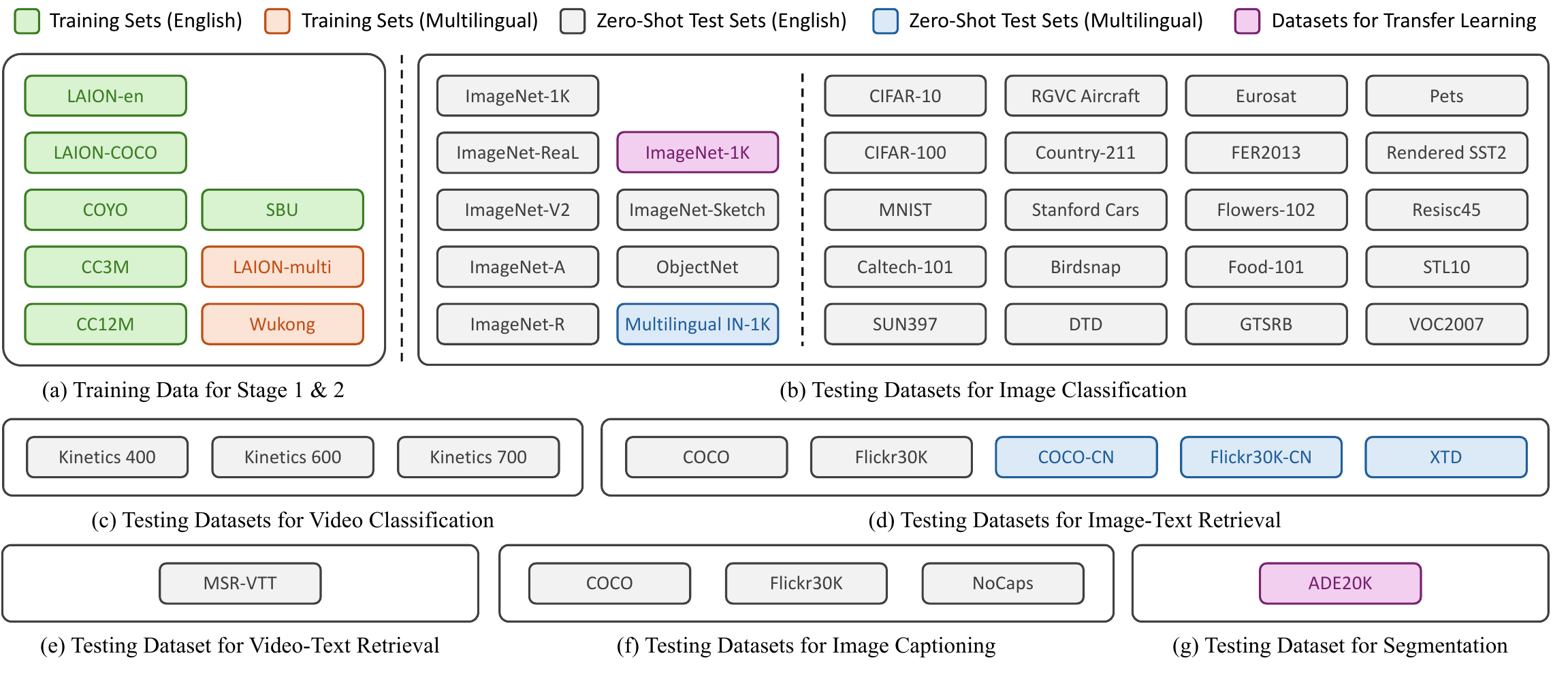}
    \caption{\textbf{Panoramic overview of the datasets used in InternVL's stage 1 and stage 2.}
    During the training of stage 1 and stage 2, we utilize web-scale image-text data from a variety of sources to train our InternVL model, as shown in (a).
    To assess InternVL's capabilities in handling generic visual-linguistic tasks, we conducted extensive validations across a range of tasks and datasets, including (b) image classification, (c) video classification, (d) image-text retrieval, (e) video-text retrieval, (f) image captioning, and (g) semantic segmentation.
    } 
    \label{fig:datasets_stage1_2}
\end{figure*}

\noindent
\textbf{Settings of ImageNet Linear Probing.}
We follow the common practices of linear probing in previous methods \cite{he2022mae,oquab2023dinov2,dehghani2023vit22b}.
Specifically, we employ an additional BatchNorm \cite{ioffe2015batch} to normalize the pre-trained backbone features during training. 
Besides, we concatenate the average-pooled patch token features with the class token. 
The linear head is trained using the SGD optimizer for 10 epochs on ImageNet-1K \cite{deng2009imagenet}, with a total batch size of 1024, a peak learning rate of 0.2, 1 epoch warm-up, and no weight decay.
Data augmentation involves random-resized-crop and flip.
For more training details, please see Table \ref{tab:train_cfg_in_linear}.

\noindent
\textbf{Settings of ADE20K Semantic Segmentation.}
In Table \ref{tab:train_cfg_ade20k}, we have listed the hyperparameters for three different configurations in ADE20K semantic segmentation, including linear probing, head tuning, and full-parameter tuning.

\subsection{Data Preparation for Pre-training}

\noindent\textbf{Training Data for Stage 1 \& Stage 2.}
During the first and second stages, we employed a vast collection of image-text pair data (see Figure \ref{fig:datasets_stage1_2} (a)), such as LAION-en \cite{schuhmann2022laion5b}, LAION-multi \cite{schuhmann2022laion5b}, LAION-COCO \cite{schuhmann2022laioncoco}, COYO \cite{byeon2022coyo}, Wukong \cite{gu2022wukong}, among others \cite{sharma2018cc3m, changpinyo2021cc12m, ordonez2011sbu}.
A detailed introduction to these datasets is provided in Table \ref{tab:data_intro_stage1_2_page1}.

\noindent
\textbf{Training Data Cleaning for Stage 1 \& Stage 2.} 
To fully utilize web-scale image-text data, we adopted different data filtering strategies in stage 1 and stage 2.

(1) Stage 1: In the first stage, we applied only minor data filtering, thus retaining the vast majority of the data.
We considered six factors: CLIP similarity, watermark probability, unsafe probability, aesthetic score, image resolution, and caption length, to remove extreme data points and avoid disrupting training stability. 
Additionally, we removed data that was duplicated with ImageNet-1K/22K \cite{deng2009imagenet}, Flickr30K \cite{plummer2015flickr30k}, and COCO \cite{lin2014microsoft} to ensure the reliability of our zero-shot evaluations.
Due to download failures and the use of our data filtering pipeline, the total amount of data retained in the first stage was 4.98 billion.

(2) Stage 2: In the second stage, we implemented a more stringent data filtering strategy. With generative supervision included, we deleted most of the low-quality data based on the captions, mainly considering the length, completeness, readability, and whether they were gibberish or boilerplate (like menus, error messages, or duplicate text), contained offensive language, placeholder text, or source code. We retained only 1.03 billion entries.

\noindent\textbf{Testing Datasets for Image Classification.}
We conducted extensive validation on image classification tasks (see Figure \ref{fig:datasets_stage1_2} (b)), including the linear probing performance of InternViT-6B and the zero-shot performance of InternVL-C. These datasets used are listed in Table \ref{tab:data_intro_stage1_2_page1}.

\noindent\textbf{Testing Datasets for Video Classification.}
As shown in Figure \ref{fig:datasets_stage1_2} (c), to evaluate the capabilities of video classification, we utilize the following Kinetics datasets: Kinetics 400 \cite{carreira2017k400}, Kinetics 600 \cite{carreira2018k600},  and Kinetics 700 \cite{carreira2019k700}.

\noindent\textbf{Testing Datasets for Image-Text Retrieval.}
We use five datasets (see Figure \ref{fig:datasets_stage1_2} (d)) to evaluate InternVL's zero-shot, multilingual image-text retrieval capabilities.
A detailed introduction to these datasets is provided in Table \ref{tab:data_intro_stage1_2_page2}.

\noindent\textbf{Testing Dataset for Video-Text Retrieval.}
As shown in Figure \ref{fig:datasets_stage1_2} (e), we use the MSR-VTT \cite{xu2016msrvtt} dataset to evaluate our InternVL in zero-shot video-text retrieval.

\noindent\textbf{Testing Dataset for Image Captioning.}
As illustrated in Figure \ref{fig:datasets_stage1_2} (f), we use three image captioning datasets to test our InternVL model.
A detailed introduction to these datasets is provided in Table \ref{tab:data_intro_stage1_2_page3}.

\noindent\textbf{Testing Dataset for Semantic Segmentation.}
We use the ADE20K \cite{zhou2017ade20k} dataset to study the pixel-level perceptual capacity of InternViT-6B, as shown in Figure \ref{fig:datasets_stage1_2} (g).
A detailed introduction to this dataset is provided in Table \ref{tab:data_intro_stage1_2_page3}.

\subsection{Data Preparation for SFT}

\noindent\textbf{Training Data for SFT.}
In this stage, we collect a wide range of high-quality instruction data.
For non-dialogue datasets, we follow the method described in \cite{liu2023improved} for conversion.
A detailed introduction is provided in Table \ref{tab:data_intro_stage3_page1}.

\noindent\textbf{Testing Datasets for SFT.}
We validate the effectiveness of our supervised fine-tuned InternVL-Chat models on three tasks, including image captioning, visual question answering, and multi-modal dialogue. 
There datasets are listed in Table \ref{tab:data_intro_stage3_page2}.
For most of these datasets, we employ the same response formatting prompt as for LLaVA-1.5~\cite{liu2023improved}.

\newpage

\begin{table*}[t]
\small
\renewcommand{\arraystretch}{1.1}
    \centering

    \setlength\tabcolsep{5pt}
    \begin{tabular}{p{3.2cm}|p{13.5cm}}
        dataset & introduction  \\
        \hline

    \multicolumn{2}{l}{\emph{Training Data for Stage 1 \& Stage 2.}}    \\ 
    
    LAION-en~\cite{schuhmann2022laion5b} &  LAION-en is a part of the LAION-5B dataset, containing 2.32 billion English-only image-text pairs.  \\
    
    \rowcolor{gray!15}
    LAION-multi~\cite{schuhmann2022laion5b} & LAION-multi is another segment of LAION-5B, featuring 2.26 billion image-text pairs across more than 100 languages, and is ideal for multilingual studies. \\
    
    Laion-COCO~\cite{schuhmann2022laioncoco} & Laion-COCO comprises 663 million synthetic captions for web images, generated using a blend of BLIP-L/14 \cite{li2022blip} and CLIP models \cite{radford2021clip}.  \\
    
    \rowcolor{gray!15}
    COYO~\cite{byeon2022coyo} &  COYO-700M is a large-scale dataset that contains 747 million image-text pairs as well as many other meta-attributes to increase the usability to train various models. It follows a similar strategy to previous vision-language datasets, collecting many informative pairs of alt-text and its associated image in HTML documents.  \\
    
    Wukong~\cite{gu2022wukong} & Wukong is a large-scale Chinese image-text dataset for benchmarking different multi-modal pre-training methods. It contains 100 million Chinese image-text pairs from the web. \\

    \rowcolor{gray!15}
    CC3M~\cite{sharma2018cc3m} & This dataset consists of approximately 3 million images, each annotated with a caption.  \\
    
    CC12M~\cite{changpinyo2021cc12m} &  CC12M is a dataset with 12 million image-text pairs. It is larger and covers a much more diverse set of visual concepts than the CC3M \cite{sharma2018cc3m}.  \\

    \rowcolor{gray!15}
    SBU~\cite{ordonez2011sbu} &  The SBU Captioned Photo Dataset is a collection of over 1 million images with associated text descriptions extracted from Flicker. \\

    \hline
    \multicolumn{2}{l}{\emph{Testing Datasets for Image Classification.}}    \\ 

    ImageNet-1K \cite{deng2009imagenet} & A large-scale dataset commonly used in image classification, consisting of over 1 million images across 1K different classes. \\

    \rowcolor{gray!15}
    ImageNet-ReaL \cite{beyer2020imagenetreal} & It contains ImageNet val images augmented with a new set of ``re-assessed" labels. These labels are collected using an enhanced protocol, resulting in multi-label and more accurate annotations. \\
    
    ImageNet-V2 \cite{recht2019imagenetv2} & A dataset created to test the robustness of models trained on ImageNet-1K, containing new test images collected following the original methodology. \\

    \rowcolor{gray!15}
    ImageNet-A \cite{hendrycks2021imagenet_a} & It consists of real-world, unmodified, and naturally occurring examples that are misclassified by ResNet models \cite{he2016deep}. It's designed to highlight the challenges of adversarial examples in natural settings. \\
    
    ImageNet-R \cite{hendrycks2021imagenet_r} & A set of images labeled with ImageNet labels obtained by collecting art, cartoons, deviantart, graffiti, embroidery, graphics, origami, paintings, patterns, plastic objects, plush objects, sculptures, sketches, tattoos, toys, and video game renditions of ImageNet classes. It has renditions of 200 ImageNet classes resulting in 30K images.  \\

    \rowcolor{gray!15}
    ImageNet-Sketch \cite{wang2019imagenet_sketch} & It consists of 51K images, approximately 50 images for each of the ImageNet classes. It is constructed using Google Image queries with the standard class name followed by ``sketch of". \\
    
    ObjectNet \cite{barbu2019objectnet} &
    ObjectNet is a crowd-sourced test set of 50K images featuring objects in unusual poses and cluttered scenes, designed to challenge recognition performance. It includes controls for rotation, background, and viewpoint, and covers 313 object classes, with 113 overlapping with ImageNet \cite{deng2009imagenet}. \\

    \rowcolor{gray!15}
    Multilingual IN-1K \cite{laion_ai_2023_clip} & An adaptation of ImageNet-1K supporting multilingual annotations, facilitating research in cross-lingual image classification. \\
    
    CIFAR-10/100 \cite{krizhevsky2009cifar} & It comprises 60K 32$\times$32 images in 10 classes (CIFAR-10) or 100 classes (CIFAR-100). \\

    \rowcolor{gray!15}
    MNIST \cite{lecun1998mnist} & A classic dataset containing 70K 28$\times$28 gray-scale images of handwritten digits. \\

    Caltech-101 \cite{fei2004learning} &
    The dataset comprises images of objects from 101 classes and a background clutter class, each labeled with a single object. It contains about 40 to 800 images per class, totaling approximately 9K images. \\

    \rowcolor{gray!15}
    SUN397 \cite{xiao2010sun} & The SUN397 or Scene UNderstanding (SUN) is a dataset for scene recognition consisting of 397 categories with 109K images. \\

    FGVC Aircraft \cite{maji2013fgvc} & The dataset contains 10K images of aircraft, with 100 images for each of 102 different aircraft model variants, most of which are airplanes. \\

    \rowcolor{gray!15}
    Country-211 \cite{radford2021clip} & It is a dataset released by OpenAI, designed to assess the geolocation capability of visual representations. It filters the YFCC100M \cite{thomee2016yfcc100m} dataset to find 211 countries that have at least 300 photos with GPS coordinates. OpenAI built a balanced dataset with 211 categories, by sampling 200 photos for training and 100 photos for testing, for each country. \\

    Stanford Cars \cite{krause2013cars} & This dataset consists of 196 classes of cars with a total of 16K images, taken from the rear. The data is divided into almost a 50-50 train/test split with 8K training images and 8K testing images. \\

    \end{tabular}
    \caption{\textbf{Introduction of datasets used in InternVL's stage 1 and stage 2.}
    In summary, we utilize a vast amount of image-text data for pre-training and conduct comprehensive evaluation across a wide range of generic visual-linguistic tasks.
}
\label{tab:data_intro_stage1_2_page1}
\end{table*}

\begin{table*}[t]
\small
\renewcommand{\arraystretch}{1.1}
    \centering

    \setlength\tabcolsep{5pt}
    \begin{tabular}{p{3.2cm}|p{13.5cm}}
        dataset & introduction  \\
        \hline

    \multicolumn{2}{l}{\emph{Testing Datasets for Image Classification.}}    \\ 
    
    Birdsnap \cite{berg2014birdsnap} & Birdsnap is a large bird dataset consisting of 49,829 images from 500 bird species with 47,386 images used for training and 2,443 images used for testing.
    Due to broken links, we are only able to download 1,845 out of the 2,443 testing images.  \\

    \rowcolor{gray!15}
    DTD \cite{cimpoi2014d2d} & The Describable Textures Dataset (DTD) contains 5,640 texture images in the wild. They are annotated with human-centric attributes inspired by the perceptual properties of textures. \\
    
    Eurosat \cite{helber2019eurosat} & This dataset is based on Sentinel-2 satellite images covering 13 spectral bands and consisting of 10 classes with 27K labeled and geo-referenced samples. \\

    \rowcolor{gray!15}
    FER2013 \cite{goodfellow2013fer2013} &
    This dataset includes around 30K RGB facial images, categorized into seven expressions: angry, disgust, fear, happy, sad, surprise, and neutral.  \\
    
    Flowers-102 \cite{nilsback2008flowers} & It is consistent with 102 flower categories commonly occurring in the United Kingdom. Each class consists of between 40 and 258 images.\\

    \rowcolor{gray!15}
    Food-101 \cite{bossard2014food101} & The Food-101 dataset consists of 101 food categories with 750 training and 250 test images per category, making a total of 101K images.\\
    
    GTSRB \cite{stallkamp2012gtsrb} & The German Traffic Sign Recognition Benchmark (GTSRB) contains 43 classes of traffic signs, split into 39,209 training images and 12,630 test images.  \\

    \rowcolor{gray!15}
    Pets \cite{parkhi2012cats} & The Oxford-IIIT Pet Dataset is a 37-category pet dataset with roughly 200 images for each class created by the Visual Geometry Group at Oxford. \\
    
    Rendered SST2 \cite{radford2021clip} & This dataset is used to evaluate the model's capability on optical character recognition. It was generated by rendering sentences in the Standford Sentiment Treebank v2 dataset. \\
    
    \rowcolor{gray!15}
    Resisc45 \cite{coates2011stl10} & This is a dataset for remote sensing scene classification. It contains 31,500 RGB images divided into 45 scene classes, each class containing 700 images. \\
    
    STL10 \cite{nilsback2008flowers} & The STL-10 dataset, inspired by CIFAR-10~\cite{krizhevsky2009cifar}, includes 10 classes with 500 training and 800 test color images each, sized 96$\times$96 pixels.   \\

    \rowcolor{gray!15}
    VOC2007 \cite{everingham2015pascal} &
    The Pascal VOC 2007 dataset focuses on recognizing objects in realistic scenarios and contains 20 object classes across 9,963 images with 24,640 labeled objects. The data has been divided into 50\% for training/validation and 50\% for testing. Following common practice, we conduct zero-shot image classification by cropping images to isolate objects using bounding boxes.\\
    
    \hline
    \multicolumn{2}{l}{\emph{Testing Datasets for Video Classification.}}    \\ 
    
    Kinetics 400 \cite{carreira2017k400} & A large-scale dataset containing around 400 human action classes with at least 400 video clips for each class, sourced from YouTube. \\

    \rowcolor{gray!15}
    Kinetics 600 \cite{carreira2018k600} & An expansion of Kinetics 400, this dataset includes 600 action classes and provides an increased diversity in video representation.  \\

    Kinetics 700 \cite{carreira2019k700} & The latest in the series, Kinetics 700 offers an even broader range with 700 action categories, further challenging the robustness of retrieval models.  \\
    
    \hline
    \multicolumn{2}{l}{\emph{Testing Datasets for Image-Text Retrieval.}}    \\ 

    COCO \cite{chen2015cococaption} & The COCO Caption dataset contains diverse images with detailed captions, widely used for image-text retrieval and image captioning tasks. \\

    \rowcolor{gray!15}
    COCO-CN \cite{li2019cococn} & COCO-CN is a bilingual image description dataset enriching COCO with manually written Chinese sentences and tags. The new dataset can be used for multiple tasks including image tagging, captioning, and retrieval, all in a cross-lingual setting. \\
    
    Flickr30K \cite{plummer2015flickr30k} & This dataset comprises 31,000 images sourced from Flickr, each annotated with five captions, making it suitable for image-text retrieval. \\

    \rowcolor{gray!15}
    Flickr30K-CN \cite{lan2017flickrcn} & Flickr30K-CN offers Chinese captions for the images, enabling studies in cross-lingual and multi-modal retrieval tasks. \\
    
    XTD \cite{aggarwal2020xtd} & A newly developed 1K multilingual test set, featuring COCO images annotated in various languages.  \\

    \hline
    \multicolumn{2}{l}{\emph{Testing Dataset for Video-Text Retrieval.}}    \\ 

    MSR-VTT \cite{xu2016msrvtt} &  This is a large-scale dataset for open-domain video captioning and video-text retrieval, comprising 10,000 video clips across 20 categories. Each clip is annotated with 20 English sentences, totaling about 29,000 distinct words in all captions. The standard division of the dataset allocates 6,513 clips for training, 497 for validation, and 2,990 for testing purposes. \\
    
    \end{tabular}
    \caption{\textbf{Introduction of datasets used in InternVL's stage 1 and stage 2.} 
    In summary, we utilize a vast amount of image-text data for pre-training and conduct comprehensive evaluation across a wide range of generic visual-linguistic tasks.
}
\label{tab:data_intro_stage1_2_page2}
\end{table*}

\begin{table*}[t]
\small
\renewcommand{\arraystretch}{1.1}
    \centering

    \setlength\tabcolsep{5pt}
    \begin{tabular}{p{3.2cm}|p{13.5cm}}
        dataset & introduction  \\
        \hline

    \multicolumn{2}{l}{\emph{Testing Datasets for Image Captioning.}}    \\ 
    
    COCO \cite{chen2015cococaption} & We use the Karpathy test set for testing. \\

    \rowcolor{gray!15}
    Flickr30K \cite{plummer2015flickr30k} & We use the Karpathy test set for testing. \\ 
    
    NoCaps  \cite{agrawal2019nocaps} & NoCaps stands out for testing models' capabilities in open-ended caption generation, using images that go beyond the training data's domain. We report the performance on the NoCaps val set. \\
    
    \hline
    \multicolumn{2}{l}{\emph{Testing Dataset for Semantic Segmentation.}}    \\ 

    ADE20K \cite{zhou2017ade20k} & ADE20K contains more than 20K scene-centric images exhaustively annotated with pixel-level objects and object parts labels. There are a total of 150 semantic categories, which include stuffs like sky, road, grass, and discrete objects like person, car, bed. We report the performance on the ADE20K val set. \\
    
    \end{tabular}
    \caption{\textbf{Introduction of datasets used in InternVL's stage 1 and stage 2.} 
    In summary, we utilize a vast amount of image-text data for pre-training and conduct comprehensive evaluation across a wide range of generic visual-linguistic tasks.
}
\label{tab:data_intro_stage1_2_page3}
\end{table*}

\begin{table*}[t]
\small
\renewcommand{\arraystretch}{1.15}
    \centering

    \setlength\tabcolsep{5pt}
    \begin{tabular}{p{3.2cm}|p{13.5cm}}
        dataset & introduction  \\
        \hline

    \multicolumn{2}{l}{\emph{Training Data for SFT.}}    \\ 
    
    COCO Caption \cite{chen2015cococaption} & It contains over 0.5 million captions describing over 110K images. Following common practice, we use the Karpathy training set for training. We transform it into a dialogue dataset using the response formatting prompt: ``Provide a one-sentence caption for the provided image." \\

    \rowcolor{gray!15}
    TextCaps \cite{sidorov2020textcaps} & TextCaps contains 145K captions for 28K images. It challenges a model to recognize text, relate it to its visual context, and decide what part of the text to copy or paraphrase. OCR tokens are used during training.  We transform it into a dialogue dataset using the response formatting prompt: ``Provide a one-sentence caption for the provided image." \\
    
    VQAv2 \cite{goyal2017vqav2} & VQAv2, the second version of the VQA dataset, features open-ended questions related to images. Answering these questions demands a grasp of vision, language, and common sense.  We convert it into a dialogue dataset using the prompt: ``Answer the question using a single word or phrase."\\

    \rowcolor{gray!15}
    OKVQA \cite{marino2019okvqa} & A dataset with over 14K questions requiring external knowledge for answers, focusing on knowledge-based visual question answering. We transform it into a dialogue dataset using the response formatting prompt: ``Answer the question using a single word or phrase." \\

    A-OKVQA \cite{schwenk2022aokvqa} & An augmented successor of OKVQA \cite{marino2019okvqa} and contains 25K questions requiring a broad base of commonsense and world knowledge to answer.  We transform it into a dialogue dataset using the response formatting prompt: ``Answer with the option’s letter from the given choices directly." \\

    \rowcolor{gray!15}
    IconQA \cite{lu2021iconqa} & A dataset with 107K questions across three sub-tasks, focusing on abstract diagram recognition and comprehensive visual reasoning. We convert it into a dialogue dataset using these prompts: ``Answer with the option’s letter from the given choices directly." and ``Answer the question using a single word or phrase." \\
    
    AI2D \cite{kembhavi2016ai2d} & AI2D features over 5K grade school science diagrams with rich annotations and 15K multiple-choice questions for diagram understanding research. We convert it into a dialogue dataset using the prompt: ``Please answer the question based on the options mentioned before."\\

    \rowcolor{gray!15}
    GQA \cite{hudson2019gqa} & GQA is a large-scale dataset with more than 110K images and 22 million questions, combining real images with balanced question-answer pairs for visual reasoning.  We transform it into a dialogue dataset using the prompt: ``Answer the question using a single word or phrase." \\

    OCR-VQA~\cite{mishra2019ocrvqa} & The OCR-VQA dataset contains 207,572 images of book covers and more than 1 million question-answer pairs about these images. We convert it into a dialogue dataset using the response formatting prompt: ``Answer the question using a single word or phrase." \\

    \rowcolor{gray!15}
    ChartQA~\cite{masry2022chartqa} & ChartQA is a dataset for question answering about charts, focusing on visual and logical reasoning. It comprises 9.6K human-written questions and 23.1K questions generated from human-written chart summaries.  We convert it using the prompt: ``Answer the question using a single word or phrase."  \\
    
    DocVQA~\cite{clark2017docqa} & The DocVQA dataset consists of 50,000 questions defined on over 12,000 document images. We convert it into a dialogue dataset using the prompt: ``Answer the question using a single word or phrase." \\

    \rowcolor{gray!15}
    ST-VQA~\cite{biten2019stvqa} & The ST-VQA dataset contains a total of 31,791 questions over 23,038 images. The training set alone consists of 26,308 questions based on 19,027 images. We convert it into a dialogue dataset using the response formatting prompt: ``Answer the question using a single word or phrase." \\

    \end{tabular}
    \caption{\textbf{Introduction of datasets used in InternVL's stage 3.} 
    We collect a wide range of high-quality instruction data. For non-dialogue datasets, we follow the response formatting prompts described in \cite{liu2023improved} for conversion. Note that only the training set is used for training.
}
\label{tab:data_intro_stage3_page1}
\end{table*}

\begin{table*}[t]
\small
\renewcommand{\arraystretch}{1.1}
    \centering

    \setlength\tabcolsep{5pt}
    \begin{tabular}{p{3cm}|p{13.7cm}}
        dataset & introduction  \\
        \hline

    \multicolumn{2}{l}{\emph{Training Data for SFT.}}    \\

    EST-VQA~\cite{wang2020estvqa} & The EST-VQA dataset provides questions, images, and answers, but also a bounding box for each question that indicates the area of the image that informs the answer.  We convert it into a dialogue dataset using the response formatting prompt: ``Answer the question using a single word or phrase." \\

    \rowcolor{gray!15}
    InfoVQA~\cite{mathew2022infographicvqa} & This dataset includes a diverse collection of infographics with natural language questions and answers. It focuses on reasoning over document layout, textual content, graphical elements, and data visualizations. We convert it into a dialogue dataset using the prompt: ``Answer the question using a single word or phrase."  \\

    LLaVAR~\cite{zhang2023llavar} & The LLaVAR dataset advances visual instruction tuning for Large Language Models by focusing on text-rich images. It incorporates 422K images processed with OCR and 16K GPT-4 generated conversations, enhancing text-based VQA performance and human interaction capabilities in diverse scenarios. Note that, we only use the 20K high-quality data for fine-tuning of LLaVAR. \\

    \rowcolor{gray!15}
    RefCOCO \cite{yu2016refcoco, mao2016refcocog} & A mixed dataset of RefCOCO~\cite{yu2016refcoco}, RefCOCO+\cite{yu2016refcoco}, and RefCOCO-g~\cite{mao2016refcocog}. We convert it into a dialogue dataset following LLaVA-1.5 \cite{liu2023improved}. \\

    Toloka \cite{ustalov2023toloka} & The TolokaVQA dataset comprises images with associated textual questions, each marked with a bounding box indicating the visual answer. It's sourced from a licensed subset of the COCO dataset and labeled on the Toloka platform. We convert it into a dialogue dataset following LLaVA-1.5 \cite{liu2023improved}. \\

    \rowcolor{gray!15}
    LLaVA-150K~\cite{liu2023llava} & This is a set of GPT-generated multi-modal instruction-following data, constructed for visual instruction tuning and building large multi-modal models towards GPT-4 vision/language capability. It includes 158K unique language-image instruction-following samples.  \\

    SVIT \cite{zhao2023svit} & This dataset includes 3.2 million visual instruction tuning data, with 1.6M conversation QA pairs, 1.6M complex reasoning QA pairs, and 106K detailed image descriptions. It is designed to improve multi-modal performance in visual perception, reasoning, and planning. For this dataset, we merge the QA pairs from the same training image into a single conversation. \\

    \rowcolor{gray!15}
    VisDial \cite{das2017visdial} & A dataset based on the COCO images, featuring dialogues created by two Amazon Mechanical Turk workers. One plays the `questioner', seeing only an image's text description, and the other, the `answerer', sees the image. They engage in a 10-round Q\&A session about the image. \\

    LRV-Instruction \cite{liu2023lrv-instruction} & 
    The LRV-Instruction dataset is designed to combat hallucination in large multi-modal models. It comprises 120K GPT-4-generated visual instructions for 16 vision-and-language tasks, including both positive and negative instructions for robust tuning. Negative instructions focus on Nonexistent and Existent Element Manipulation. This dataset helps improve accuracy and consistency in multi-modal tasks. \\

    \rowcolor{gray!15}
    LLaVA-Mix-665K \cite{liu2023improved} & LLaVA-Mix-665K is an instruction-following dataset mixed from 10 academically oriented datasets. \\

    \hline
    \multicolumn{2}{l}{\emph{Testing Dataset for SFT (Image Captioning).}}    \\ 
    COCO \cite{chen2015cococaption} & Karpathy test set is used for testing. The prompt is: ``Provide a one-sentence caption for the provided image.'' \\

    \rowcolor{gray!15}
    Flickr30K \cite{plummer2015flickr30k} & Karpathy test set is used for testing. The prompt is: ``Provide a one-sentence caption for the provided image.''  \\ 
    
    NoCaps  \cite{agrawal2019nocaps} & NoCaps val set is used for testing. The prompt is: ``Provide a one-sentence caption for the provided image.''  \\

    \hline
    \multicolumn{2}{l}{\emph{Testing Dataset for SFT (Visual Question Answering).}}    \\ 
    VQAv2~\cite{goyal2017vqav2} & VQAv2 test-dev set is used for testing. The prompt is: ``Answer the question using a single word or phrase.'' \\

    \rowcolor{gray!15}
    GQA~\cite{hudson2019gqa} & GQA test-balanced set is used. The prompt is: ``Answer the question using a single word or phrase.'' \\
    
    VizWiz~\cite{gurari2018vizwiz} & VizWiz test-dev set is used for testing. The prompt is: ``When the provided information is insufficient, respond with `Unanswerable'. Answer the question using a single word or phrase.'' \\

    \rowcolor{gray!15}
    TextVQA~\cite{singh2019textvqa} & TextVQA val set is used for testing. The prompt is: ``Answer the question using a single word or phrase.'' \\

    \hline
    \multicolumn{2}{l}{\emph{Testing Dataset for SFT (Multi-Modal Dialogue).}}    \\ 
    MME~\cite{fu2023mme} & MME is a comprehensive evaluation benchmark for multi-modal large language models. It measures both perception and cognition abilities on a total of 14 subtasks, including existence, count, position, color, poster, celebrity, scene, landmark, artwork, OCR, commonsense reasoning, numerical calculation, text translation, and code reasoning. The prompt for this dataset is: ``Answer the question using a single word or phrase.'' \\

    \rowcolor{gray!15}
    POPE~\cite{li2023pope} & POPE is a popular dataset used to evaluate object hallucination. The response formatting prompt used for this dataset is: ``Answer the question using a single word or phrase.'' \\

    \end{tabular}
    \caption{\textbf{Introduction of datasets used in InternVL's stage 3.}
    We collect a wide range of high-quality instruction data. For non-dialogue datasets, we follow the response formatting prompts described in \cite{liu2023improved} for conversion. Note that only the training set is used for training.
    We evaluate our InternVL-Chat models on three tasks, including image captioning, VQA, and multi-modal dialogue.
    For these datasets, we employ the same response formatting prompts as for LLaVA-1.5~\cite{liu2023improved}.
}
\label{tab:data_intro_stage3_page2}
\end{table*}

\newpage



\clearpage

{
    \small
    \bibliographystyle{ieeenat_fullname}
    \bibliography{main}
}

\end{document}